\documentclass{article}

\usepackage{microtype}
\usepackage{graphicx}
\usepackage{subfigure}
\usepackage{booktabs}
\usepackage{lipsum}
\usepackage{orcidlink}

\usepackage{hyperref}
\newcommand\scalemath[2]{\scalebox{#1}{\mbox{\ensuremath{\displaystyle #2}}}}

\usepackage[accepted]{icml2025}

\usepackage{amsmath}
\usepackage{amssymb}
\usepackage{mathtools}
\usepackage{amsthm}

\usepackage[capitalize,noabbrev]{cleveref}

\DeclareMathOperator*{\argmin}{argmin}

\usepackage[capitalize,noabbrev]{cleveref}

\theoremstyle{plain}
\newtheorem{theorem}{Theorem}[section]

\newtheorem{lemma}[theorem]{Lemma}

\theoremstyle{definition}
\newtheorem{definition}[theorem]{Definition}

\theoremstyle{remark}

\usepackage[disable,textsize=tiny]{todonotes}

\usetikzlibrary{
    shapes.geometric,
    positioning,
    arrows.meta,
    calc
}

\icmltitlerunning{Differential Privacy: Gradient Leakage Attacks in Federated Learning Environments}

\begin{document}

\twocolumn[
\icmltitle{Differential Privacy: Gradient Leakage Attacks in Federated\\Learning Environments}

\icmlsetsymbol{equal}{*}

\begin{icmlauthorlist}
\icmlauthor{Miguel Fernandez-de-Retana}{bcam,deusto}
\icmlauthor{Unai Zulaika}{deusto}
\icmlauthor{Rubén Sánchez-Corcuera}{deusto}
\icmlauthor{Aitor Almeida}{deusto}
\end{icmlauthorlist}

\icmlaffiliation{bcam}{\textit{Modeling \& Simulation in Life and Materials Sciences}, Basque Center for Applied Mathematics (BCAM)}
\icmlaffiliation{deusto}{\textit{Faculty of Engineering}, University of Deusto (Bilbao, Spain)}

\icmlcorrespondingauthor{Miguel Fernandez-de-Retana}{m.fernandezderetana@deusto.es}

\icmlkeywords{Data Leakage, Differential Privacy, Federated Learning, Input Reconstruction, ML Privacy}

\vskip 0.3in
]

\printAffiliationsAndNotice{}  

\begin{abstract}
Federated Learning (FL) allows for the training of Machine Learning models in a collaborative manner without the need to share sensitive data. However, it remains vulnerable to Gradient Leakage Attacks (GLAs), which can reveal private information from the shared model updates. In this work, we investigate the effectiveness of Differential Privacy (DP) mechanisms --- specifically, DP-SGD and a variant based on explicit regularization (PDP-SGD) --- as defenses against GLAs. To this end, we evaluate the performance of several computer vision models trained under varying privacy levels on a simple classification task, and then analyze the quality of private data reconstructions obtained from the intercepted gradients in a simulated FL environment. Our results demonstrate that DP-SGD significantly mitigates the risk of gradient leakage attacks, albeit with a moderate trade-off in model utility. In contrast, PDP-SGD maintains strong classification performance but proves ineffective as a practical defense against reconstruction attacks. These findings highlight the importance of empirically evaluating privacy mechanisms beyond their theoretical guarantees, particularly in distributed learning scenarios where information leakage may represent an unassumable critical threat to data security and privacy.
\end{abstract}

\textbf{Keywords:} Data Leakage $\cdot$ Differential Privacy $\cdot$ Federated Learning $\cdot$ Input Reconstruction $\cdot$ ML Privacy

\section{Introduction}

\textbf{Federated Learning} (FL) has recently emerged as a crucial paradigm in \emph{distributed} training, enabling multiple entities (or \emph{nodes}) to collaboratively train Machine Learning (ML) models without the need to share their local data. This approach is especially valuable in scenarios where data privacy is of utmost importance, such as in health-related applications. By keeping data on local nodes and only sharing updates through the exchange of gradients, FL seeks to mitigate the risks associated with exposing \emph{sensitive data}. This strategy not only reduces the need to transfer large volumes of data but also minimizes the risk of privacy breaches, since data never leaves the local host and model updates are securely aggregated on a central server. This allows models to learn from a wide variety of data without compromising individual privacy, as the updates do not directly reveal information about the local data of each node. In this sense, FL presents itself as a promising solution to address the challenges of privacy and security in ML, e.g., enabling organizations to collaborate in developing robust shared models without compromising the confidentiality of their sensitive data.

However, despite its inherent privacy advantages, FL is not immune to vulnerabilities. One of its most prominent threats are \textbf{Gradient Leakage Attacks} (GLAs), in which an adversary --- potentially a malicious participant or an entity intercepting communications as a \emph{man-in-the-middle} --- can infer sensitive information about a client's private training data by analyzing the gradients, or parameter updates, shared with the central server \cite{zhu2019deep, zhao2020idlg}. It has been shown that, under certain conditions, original data instances can be reconstructed with high fidelity from these gradients \cite{gradient_leakage_attacks}, posing a considerable privacy breach and, thus, a significant risk to data security.

To counter these threats and strengthen the privacy guarantees of FL, \textbf{Differential Privacy} (DP) has become the gold standard \cite{dwork2006our, dwork2014algorithmic}. DP provides a rigorous framework for quantifying and limiting the information that an algorithm's output reveals about any individual record. In the context of ML, and more specifically FL, this is typically achieved through mechanisms such as Differentially-Private Stochastic Gradient Descent (DP-SGD) \cite{DP_SGD}, which introduce calibrated noise into the training process (specifically, by clipping the norm of individual gradients and adding Gaussian noise to the aggregated sum) to obscure the contribution of each instance. Nevertheless, while effective in guaranteeing privacy, the application of DP often entails a \emph{trade-off} in model performance (e.g., in reduced accuracy) and may require careful tuning of its hyperparameters (such as the privacy budget $\varepsilon$ and the failure probability $\delta$) \cite{DP_SGD}. Moreover, most recently, alternatives and refinements to DP-SGD have emerged, such as regularization-based approaches \cite{loss_function_regularization}, which aim to offer a different balance between privacy, utility, and computational efficiency.

This work explores the intersection of federated learning, gradient leakage attacks, and differential privacy. Our main objective is to investigate and evaluate the effectiveness of DP mechanisms, particularly DP-SGD and related approaches, as defenses against GLAs in a simulated FL environment. To this end, we first introduce the fundamental concepts of FL, GLAs, and DP. We then compare the performance of several computer vision models trained with, and without, differential privacy protection on a moderately complex classification task. Then, we simulate a representative example of a gradient leakage attack and assess, both quantitatively and qualitatively, the impact of DP on the quality of the reconstructed data. Finally, we discuss the results, highlighting the inherent trade-off between privacy and utility, and outline potential limitations in this critical area of information security within distributed ML.

In summary, the main contributions of our work\footnote{Code available at \href{https://github.com/miguelfrndz/Differential-Privacy-GL-Attacks}{https://github.com/miguelfrndz/Differential-Privacy-GL-Attacks}} are the following:
\begin{itemize}
    \item To present the \emph{theoretical foundations} behind differential privacy and gradient leakage attacks.
    \item To compare the performance of models trained with and without \emph{privacy protection} on a moderately complex classification task.
    \item To simulate a representative example of a \emph{gradient leakage attack} and investigate the impact of differential privacy on the quality of the reconstructed data.
\end{itemize}

\section{Related Work}

The study of security and privacy in distributed Machine Learning builds upon several key concepts. Below, we provide a brief snapshot of the \textit{state-of-the-art} in the areas most relevant to this work, with technical details to be further elaborated later in Section \ref{sec:background}.

\textbf{Federated Learning (FL):} Originally introduced by \citet{mcmahan2017communication} and further developed in subsequent works such as those by \citet{kairouz2021advances} and \citet{li2020federated}, Federated Learning is a decentralized ML paradigm that allows multiple clients to collaboratively train a \emph{shared model} without the need to exchange their local data. This distributed approach is particularly well-suited for privacy-sensitive applications, such as those involving medical or biometric information, since the raw data remains on the user devices or within the infrastructure of participating organizations. FL aims to leverage the computational resources and data heterogeneity distributed across clients, while maintaining a foundational level of local data privacy and security.

\textbf{Gradient Leakage Attacks (GLA):} However, it soon became evident that the \textit{de facto} privacy offered by standard FL was limited against certain adversaries. Recent studies have demonstrated that an attacker with access to gradients, or parameter updates shared during federated training, can reconstruct, with surprising fidelity in some cases, the private data samples used to compute those gradients \cite{geiping2020inverting, huang2021evaluating, zhu2019deep, zhao2020idlg}. These attacks, generically referred to as Gradient Leakage Attacks (GLA), expose a critical vulnerability that can compromise the confidentiality of the obfuscated data. 

\textbf{Differential Privacy (DP):} Introduced by \citet{dwork2006our} and formalized in subsequent works such as \cite{dwork2014algorithmic}, Differential Privacy has become the reference framework for providing robust and quantifiable privacy guarantees in data analysis and Machine Learning. DP offers protection against a wide range of inference attacks, including GLAs, by mathematically ensuring that the output of an algorithm (e.g., model parameters or aggregated gradients) does not reveal excessive information about any individual entry in the original dataset. In Machine Learning, DP is commonly implemented through mechanisms such as Differentially-Private Stochastic Gradient Descent \textbf{(DP-SGD)} \cite{DP_SGD}, or via explicit \textbf{regularization methods} \cite{loss_function_regularization}, which introduce calibrated noise during training to obscure individual contributions, although this often entails a \emph{trade-off} in the final model quality. Moreover, more recent approaches explore alternative pathways, such as generating \emph{synthetic datasets} that intrinsically comply with differential privacy, e.g., by leveraging Large Language Models (LLMs) to create private synthetic textual data \cite{synthetic_private_data}.

This work lies at the intersection of these three areas, specifically evaluating the effectiveness of Differential Privacy as a countermeasure against Gradient Leakage Attacks in Federated Learning environments. However, it is worth noting that, although the context of this work is framed within federated learning, we will not delve deeply into its specific aspects. Instead, we assume that model gradients have been intercepted in an FL setting and focus \emph{exclusively} on the issues related to DP and the aforementioned attacks.

\section{Background}\label{sec:background}

In this section, we detail the essential theoretical foundations related to the concepts of Federated Learning (Section \ref{sec:FL}), Gradient Leakage Attacks (Section \ref{sec:GLA}), and Differential Privacy (Section \ref{sec:DP}), providing the necessary groundwork to understand the analysis and results presented later.

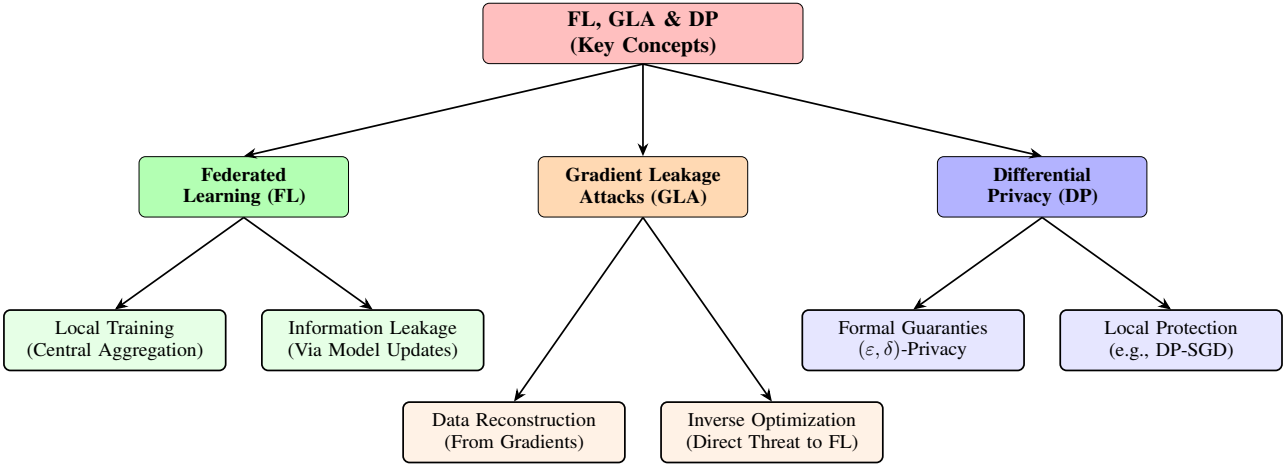
\begin{figure*}[!t]
    \centering
    \resizebox{\linewidth}{!}{
    \begin{tikzpicture}[
        every node/.style={
            draw,                
            rectangle,           
            rounded corners=3pt, 
            align=center,        
            text width=3.4cm,    
            minimum height=2.8em,
            font=\small,
            inner sep=3pt
        },
        level 1/.style={
            sibling distance=6.5cm,
            level distance=2.5cm    
            },
        level 2/.style={
            sibling distance=4.2cm, 
            level distance=2.5cm    
            },
        edge from parent path={(\tikzparentnode.south) -- (\tikzchildnode.north)}, 
        edge from parent/.style={draw, thick, -{Stealth[length=6pt, width=5pt]}}, 
        root/.style={fill=red!25, font=\bfseries, text width=5cm}, 
        fl/.style={fill=green!30, text width=3.2cm},
        fl_leaf/.style={fill=green!10},
        gla/.style={fill=orange!30, text width=3.2cm},
        gla_leaf/.style={fill=orange!10}, 
        dp/.style={fill=blue!30, text width=3.2cm},
        dp_leaf/.style={fill=blue!10},
        gla_leaf_shifted/.style={
            fill=orange!10,          
            yshift=-1.5cm             
        },
        grow=down               
    ]

    \node [root] {FL, GLA \& DP\\(Key Concepts)}
        child { node [fl] {\textbf{Federated Learning (FL)}}
            child { node [fl_leaf] {Local Training \\ (Central Aggregation)} }
            child { node [fl_leaf] {Information Leakage \\ (Via Model Updates)} }
        }
        child { node [gla] {\textbf{Gradient Leakage\\Attacks (GLA)}}
            child { node [gla_leaf_shifted] {Data Reconstruction \\ (From Gradients)} } 
            child { node [gla_leaf_shifted] {Inverse Optimization \\ (Direct Threat to FL)} } 
        }
        child { node [dp] {\textbf{Differential\\Privacy (DP)}}
            child { node [dp_leaf] {Formal Guaranties \\ $(\varepsilon, \delta)$-Privacy} }
            child { node [dp_leaf] {Local Protection \\ (e.g., DP-SGD)} }
        };

    \end{tikzpicture}
    }
    \caption{Conceptual Summary of the Three Pillars of This Work: Federated Learning (FL), Gradient Leakage Attacks (GLA), and Differential Privacy (DP)}
\end{figure*}

\subsection{Federated Learning (FL)}\label{sec:FL}

\textbf{Federated Learning (FL)} is a distributed Machine Learning paradigm designed to enable multiple \emph{clients} (e.g., mobile devices, institutions, local servers) to collaboratively train a global model without the need to share their local training data \cite{mcmahan2017communication, kairouz2021advances, li2020federated}. This allows for addressing critical concerns related to privacy, security, and communication costs typically associated with centralizing data in traditional ML workflows. Additionally, in the context of increasing regulatory constraints on data protection and governance (e.g., GDPR, HIPAA), FL offers a viable solution for complying with legal and ethical requirements imposed by current legislation.

Typically, a standard FL system consists of the following two main components:
\begin{itemize}
    \item A set of \textbf{client nodes} (or \textit{workers}), denoted by $k \in \{1, \dots, N\}$, each with its own local dataset $D_k$, which is not shared with other clients or with the server.
    \item A \textbf{central server} (or \textit{coordinator}) that orchestrates the training process and maintains the state of the global model.
\end{itemize}

The training process itself is iterative and generally unfolds over multiple \textit{communication rounds}. In each round $t$:
\begin{enumerate}
    \item \textbf{Distribution:} The central server selects a subset of available clients (e.g., $S_t \subseteq \{1, \dots, N\}$) and sends them the current parameters of the global model $\theta_t$.
    \item \textbf{Local Training:} Each client $k \in S_t$ performs a \emph{local training} using the received model $\theta_t$ and its own dataset $D_k$. This local training usually consists of running one or more steps (epochs) of a standard optimization algorithm, such as Stochastic Gradient Descent (SGD), to minimize a local loss function $\mathcal{L}(\theta_t^{(k)}; D_k)$. 
    
    The result is an updated set of local parameters $\theta_{t+1}^{(k)}$, or alternatively, a ``\textit{computed}'' update such as the difference $\Delta \theta_t^{(k)}$ or the average gradient $\nabla \mathcal{L}(\theta_t^{(k)}; D_k)$.
    
    \begin{equation}\label{eq:local_param_update}
        \theta_{t+1}^{(k)} = \theta_t - \eta^{(k)} \nabla \mathcal{L}(\theta_t^{(k)}; D_k)
    \end{equation}
    
    \item \textbf{Communication:} Clients $k \in S_t$ send their local updates (i.e., $\theta_{t+1}^{(k)}$ or $\Delta \theta_t^{(k)}$) back to the central server. It is important to note that the raw data $D_k$ \textit{never} leaves the client during this process.
    \item \textbf{Aggregation:} The central server collects the updates received from the clients participating in round $t$ and \emph{aggregates} them to produce an improved global model $\theta_{t+1}$. Several aggregation strategies exist, the most common being \textit{Federated Averaging (FedAvg)} \cite{mcmahan2017communication}. In FedAvg, assuming clients return their local parameters $\theta_{t+1}^{(k)}$, the server computes the new global model as a weighted average as in \eqref{eq:fed_avg}.
    
    \begin{equation}\label{eq:fed_avg}
        \theta_{t+1} = \sum_{k \in S_t} \frac{n_k}{N_t} \theta_{t+1}^{(k)}
    \end{equation}
    
    where $n_k = |D_k|$ is the number of data samples on client $k$, and $N_t = \sum_{k \in S_t} n_k$ is the total number of samples across the participating clients in round $t$. This weighting ensures that clients with more data have a greater influence on the global model. The new global model $\theta_{t+1}$ then becomes the reference for the next training round.

    Other aggregation alternatives include:
    \begin{itemize}
        \item Variants of FedAvg with different weighting schemes or adaptive learning rates.
        \item Algorithms that aggregate gradients or parameter differences instead of absolute parameters.
        \item Use of \emph{Secure Aggregation} protocols \cite{bonawitz2017practical}, which employ cryptographic techniques to allow the server to compute the sum (or weighted average) of updates without accessing the individual updates $\theta_{t+1}^{(k)}$ or $\Delta \theta_t^{(k)}$ from each client. This provides an extra layer of privacy, protecting the updates even from the central server itself.
    \end{itemize}

\end{enumerate}

This four-step cycle is repeated for a predefined number of rounds or until the global model $\theta_t$ converges according to some criterion (e.g., accuracy on a global validation set or the stabilization of the loss function).

While the federated learning paradigm improves privacy by avoiding the need to centralize raw data, the updates shared by clients may still \textbf{reveal sensitive information} about their local data. Therefore, it is essential to implement additional privacy and security mechanisms to protect data confidentiality throughout the federated training process.

\subsection{Gradient Leakage Attacks (GLA)}\label{sec:GLA}

Although Federated Learning proposes a framework that avoids sharing raw local training data, the model updates (i.e., typically the gradients) that clients send to the central server are not free from privacy concerns. It has been shown that these updates, while seemingly obfuscated, can contain enough information for an adversary to infer, or even reconstruct with relative fidelity, the private data samples used to compute them \cite{zhu2019deep, zhao2020idlg, geiping2020inverting, gong2023gradient}. These methods are collectively known as \textbf{Gradient Leakage Attacks (GLA)}, and we provide below a brief description of how they work.

A typical GLA scenario assumes an adversary with specific capabilities. Generally, a \emph{white-box} attack is considered, where the attacker knows the model architecture, the loss function $\mathcal{L}$, and the model parameters $\theta_t$ prior to the client's local update. The goal of the attacker is to \emph{intercept} the local update from client $k$ before it is securely aggregated with the updates from other clients. This adversary could be an external entity intercepting communications (i.e., a \textit{man-in-the-middle} attack), or in some cases, the central server itself if it has been compromised and no secure aggregation has been implemented \cite{bonawitz2017practical}, or even compromised client nodes acting maliciously \cite{gong2023gradient}.

The core idea behind GLAs is to formulate data reconstruction as an \emph{inverse optimization problem}. The attacker initializes a ``\textit{fake}'' or ``\textit{seed}'' input $(x_{\text{rec}}^{(0)}, y_{\text{rec}}^{(0)})$ with random values (e.g., Gaussian noise for $x_{\text{rec}}^{(0)}$ and a random or inferred label for $y_{\text{rec}}^{(0)}$). The attacker then computes the gradient $\nabla \theta_{\text{att}}^{(0)}$ that this fake input would produce in the model $f$ with the known parameters $\theta_t$. The goal is to iteratively adjust $(x_{\text{rec}}, y_{\text{rec}})$ to minimize the discrepancy between the calculated gradient $\nabla \theta_{\text{att}}$ and the \textit{real} intercepted gradient $\nabla \theta_t^{(k)}$ as in \eqref{eq:gla_objective}.
\begin{equation}\label{eq:gla_objective}
(x_{\text{rec}}^*, y_{\text{rec}}^*) = \argmin_{x', y'} \|\nabla_{\theta_t} \mathcal{L}(f(x'; \theta_t), y') - \nabla \theta_t^{(k)} \|^2_2
\end{equation}
where $\|\cdot\|_2^2$ denotes the squared Euclidean norm.

If the attacker intercepts the updated parameters $\theta_{t+1}^{(k)}$ instead of the gradient directly, they can estimate the original gradient, assuming they know, or can accurately estimate, the local learning rate $\eta^{(k)}$ used by the client:
\begin{equation}
\nabla \theta_t^{(k)} \approx \frac{\theta_{t+1}^{(k)} - \theta_t}{\eta^{(k)}}
\end{equation}

The optimization in \eqref{eq:gla_objective} is typically carried out using gradient-based methods, where the update is applied directly to the variables $x_{\text{rec}}$ and $y_{\text{rec}}$. However, optimizing over $y_{\text{rec}}$ (which often belongs to a discrete space) can be challenging. A major breakthrough, proposed by \citet{zhao2020idlg}, showed that the correct label $y_{\text{rec}}$ can often be \emph{inferred directly} by analyzing the signs or magnitudes of the components of the intercepted gradient $\nabla \theta_t^{(k)}$ corresponding to the weights in the final layer of the model. This allows $y_{\text{rec}}$ to be fixed, simplifying the optimization problem to focus solely on $x_{\text{rec}}$.

Algorithm \ref{alg:gradient_reconstruction} summarizes the general process of a gradient-based reconstruction attack. After obtaining the target gradient $\nabla \theta_t^{(k)}$ (either directly or by deriving it from $\theta_{t+1}^{(k)}$), a seed $x_{\text{rec}}^{(0)}$ is initialized, and a label $y_{\text{rec}}$ is either inferred or initialized. Then, in an iterative loop governed by a stopping condition $T$:
\begin{enumerate}
    \item The attacker computes the gradient $\nabla \theta_{\text{att}}^{(\tau)}$ using the current seed $(x_{\text{rec}}^{(\tau)}, y_{\text{rec}})$.
    \item A reconstruction loss $D^{(\tau)}$ is computed to measure the distance between $\nabla \theta_{\text{att}}^{(\tau)}$ and $\nabla \theta_t^{(k)}$, potentially including \emph{regularization terms}, weighted by $\alpha$, such as the distance between the model output $f(x_{\text{rec}}^{(\tau)}; \theta_t)$ and the label $y_{\text{rec}}$ (e.g., classification loss or regression equivalent), or the total variation loss of $x_{\text{rec}}^{(\tau)}$ \cite{geiping2020inverting}.
    \item The seed $x_{\text{rec}}^{(\tau)}$ is updated by taking a gradient descent step on the loss $D^{(\tau)}$ with respect to $x_{\text{rec}}^{(\tau)}$, using a learning rate $\eta'$ for the attack. As noted earlier, the same can be done for $y_{\text{rec}}^{(\tau)}$ if it has not been directly inferred from the gradients.
\end{enumerate}
This process continues for a fixed number of iterations $T$, or until the reconstruction loss $D^{(\tau)}$ converges below a predefined threshold (both are valid termination criteria). The choice of optimizer for updating $\left(x_{\text{rec}}, y_{\text{rec}}\right)$ is important; although the algorithm shows simple gradient descent, more sophisticated optimizers such as L-BFGS \cite{liu1989limited} or Adam \cite{kingma2014adam} are commonly used in practice due to their efficiency and ability to handle high-dimensional problems \cite{gradient_leakage_attacks}. Hyperparameters such as the number of iterations $T$, the learning rate $\eta'$, and the regularization weight $\alpha$ must be carefully tuned to achieve high-quality reconstructions.

The demonstrated existence and effectiveness of GLAs underlines a fundamental vulnerability in standard Federated Learning, and strongly motivates the need for more robust privacy-preserving mechanisms, such as Differential Privacy, which we discuss in the next section.

\begin{algorithm}[t]
\caption{Gradient-Based Reconstruction Attack \cite{gradient_leakage_attacks, gong2023gradient}}
\label{alg:gradient_reconstruction}
\begin{algorithmic}[1]
\STATE \textbf{Input:} differentiable model $f(x; \theta_t)$, local gradient from node $k$ $\nabla \theta_t^{(k)}$, client learning rate $\eta^{(k)}$, stopping condition $T$, attack learning rate $\eta'$, regularization weight $\alpha$
\IF{$\theta_t^{(k)}$ updated $\theta_{t+1}^{(k)}$ is available}
    \STATE $$\nabla \theta_{t}^{(k)} \gets \dfrac{\left(\theta_{t+1}^{(k)} - \theta_t\right)}{\eta^{(k)}}$$
\ENDIF
\STATE \textit{Initialize attack seed:} $x^{(0)}_{\text{rec}}$
\STATE \textit{Infer label:} $y_{\text{rec}} \gets \argmin_{i} \left(\nabla_i \theta_t^{(k)}\right)$
\FOR{$\tau = 1$ \textbf{to} $T$}
    \STATE \textit{Compute attack gradient:} $$\nabla \theta_{\text{att}}^{(\tau)} \gets \frac{\partial \mathcal{L}(f(x_{\text{rec}}^{(\tau)}; \theta_t), y_{\text{rec}})}{\partial \theta_t}$$
    \STATE \textit{Compute reconstruction loss:} $$D^{(\tau)} \gets \|\nabla \theta_{\text{att}}^{(\tau)} - \nabla \theta_t^{(k)}\|^2 + \alpha \|f(x_{\text{rec}}^{(\tau)}; \theta_t) - y_{\text{rec}}\|^2$$
    \STATE \textit{Update seed:} $x_{\text{rec}}^{(\tau+1)} \gets x_{\text{rec}}^{(\tau)} - \eta' \dfrac{\partial D^{(\tau)}}{\partial x_{\text{rec}}^{(\tau)}}$
\ENDFOR
\STATE \textbf{Output:} Reconstructed data $(x_{\text{rec}}, y_{\text{rec}})$
\end{algorithmic}
\end{algorithm}

\subsection{Differential Privacy (DP)}\label{sec:DP}

\textbf{Differential Privacy (DP)} \cite{dwork2006our, dwork2014algorithmic} has become the \textit{de facto} standard for providing reliable and quantifiable privacy guarantees in the training of Machine Learning models, where its goal is to limit the amount of information an adversary can infer about the dataset used to train a model. We present below some basic definitions and theorems related to differential privacy. The two theorems presented, following the notation commonly used across the literature, are fundamental for understanding the concept of DP and its application in the context of information security. In this case, both theorems are stated according to the definition of \textit{differential privacy} introduced by \citet{dwork2006our} and in the seminal work of \citet{DP_SGD}. In essence, differential privacy is based on the idea that the output of an algorithm should not depend \textit{excessively} on any single individual instance, ensuring that private information cannot be inferred from the algorithm’s output.

Formally, differential privacy is defined in terms of the \textit{indistinguishability} of the outputs of a model $\mathcal{M}$ --- or more generally, an algorithm $\mathcal{A}$ --- when applied to \emph{neighboring} datasets, i.e., datasets that differ by a single record (see Definition \ref{def:vecinity_dp}). Based on this notion of neighborhood, the two main variants of differential privacy are defined in Theorems \ref{thm:eps_dp} and \ref{thm:eps_delta_dp}.

\begin{definition}[Neighborhood in $\mathcal{D}$]\label{def:vecinity_dp}
    Let $D, D' \in \mathcal{D}$ be two \textit{datasets} defined over the same domain $\mathcal{D}$. We say that $D$ and $D'$ are \textit{neighbors} or \textit{adjacent} if they differ in exactly one instance, that is, if $D$ can be obtained from $D'$ (or vice versa) by adding or removing a single element.
\end{definition}

\begin{theorem}[$\varepsilon$-Differential Privacy]\label{thm:eps_dp}
    An algorithm $\mathcal{A}\colon \mathcal{D} \to \mathcal{R}$ is $\varepsilon$-differentially private if, for every pair of neighboring datasets $D, D' \in \mathcal{D}$ and for any subset $E \subseteq \mathcal{R}$, it holds that:
    \begin{equation}
        P\left(\mathcal{A}(D)\in E\right)\leq e^\varepsilon\cdot P\left(\mathcal{A}(D')\in E\right)
    \end{equation}
\end{theorem}
That is, the probability that algorithm $\mathcal{A}$ produces an output in the set $E$ does not change significantly regardless of whether dataset $D$ or $D'$ is used.

\begin{theorem}[$(\varepsilon, \delta)$-Differential Privacy]\label{thm:eps_delta_dp}
    An algorithm $\mathcal{A}\colon \mathcal{D} \to \mathcal{R}$ is \((\varepsilon, \delta)\)-differentially private if, for every pair of neighboring datasets $D, D' \in \mathcal{D}$ and for any subset $E\subseteq\mathcal{R}$, it holds that:
    \begin{equation}
        P\left(\mathcal{A}(D)\in E\right) \leq e^\varepsilon\cdot P\left(\mathcal{A}(D')\in E\right) + \delta
    \end{equation}
\end{theorem}
Here, $\varepsilon$ is the \textbf{privacy budget}, which measures the upper bound on per-record privacy leakage. A lower $\varepsilon$ implies stronger privacy. The parameter $\delta$, typically chosen to be a small value (e.g., less than the inverse of the dataset size), represents the probability that the $\varepsilon$-DP guarantee may fail. In other words, with probability $(1-\delta)$, the algorithm behaves as $\varepsilon$-DP.

Moreover, this definition can be naturally extended to the case where the two datasets $D,D'\in \mathcal{D}$ differ in more than one instance. In this case, we talk about \emph{group privacy}, and although it is not commonly used in practice, we include its definition in Lemma \ref{lemma:group_priv} \cite{notes_DPSGD, jagielski2020auditing}.

\begin{lemma}[Group Differential Privacy]\label{lemma:group_priv}
    An algorithm $\mathcal{A}\colon \mathcal{D} \to \mathcal{R}$ is group \((\varepsilon, \delta)\)-differentially private if, for every pair of neighboring datasets $D, D' \in \mathcal{D}$ and for any subset $E\subseteq\mathcal{R}$, it holds that:
    \begin{equation}
        P\left(\mathcal{A}(D)\in E\right) \leq e^{k\varepsilon}\cdot P\left(\mathcal{A}(D')\in E\right) + \dfrac{e^{k\varepsilon}-1}{e^{\varepsilon}-1}\delta
    \end{equation}
    where \(k\) is the number of instances by which the datasets differ.
\end{lemma}

In summary, the main idea behind DP is that the inclusion, or exclusion, of certain \textit{private} training instances should not significantly affect the outcome of our model. In this setting, $\varepsilon$ represents the \textit{privacy budget}, which quantifies the level of privacy achieved, and $\delta$ corresponds to a small probability of error in maintaining or guaranteeing this privacy.

\begin{algorithm}[t]
\caption{Differentially-Private Stochastic Gradient Descent (DP-SGD)}
\label{alg:DP_SGD}
\begin{algorithmic}[1]
\STATE \textbf{Input:} loss function $\mathcal{L}(\theta)$, mini-batch size $L$, learning rate $\eta$, clipping parameter $C$, noise $\sigma$, number of iterations $T$
\STATE \textbf{Initialize:} $\theta_0$ randomly
\FOR{$t = 1$ \textbf{to} $T$}
    \STATE Sample mini-batch $\mathcal{B}_t$ of $L$ instances
    \FOR{each $x_i \in \mathcal{B}_t$}
        \STATE \textit{Compute gradient:} $g_i = \nabla_\theta \mathcal{L}(\theta_{t-1}, x_i)$
        \STATE \textit{Clip:} $\Bar{g}_i \gets g_i / \max\left(1, \frac{\|g_i\|_2}{C}\right)$
    \ENDFOR
    \STATE \textit{Compute aggregated gradient:} $$\tilde{g}_t \gets \frac{1}{L}\left(\sum_{i=1}^{L} \Bar{g}_i\right) + \mathcal{N}(0, \sigma^2 C^2 I)$$
    \STATE \textit{Update parameters:} $\theta_t \gets \theta_{t-1} - \eta \tilde{g}_t$
\ENDFOR
\STATE \textbf{Output:} final parameters $\theta_T$
\end{algorithmic}
\end{algorithm}

The most widely extended mechanism for achieving differential privacy is \textbf{Differentially-Private Stochastic Gradient Descent (DP-SGD)}, proposed by \citet{DP_SGD} and presented in Algorithm \ref{alg:DP_SGD}. DP-SGD modifies the standard SGD procedure at two key steps in each training iteration over a mini-batch $\mathcal{B}_t$ of size $L$:

\begin{enumerate}
    \item \textbf{Gradient Clipping:} Before aggregating the mini-batch gradients, the individual gradients $g_i$ for each sample $x_i \in \mathcal{B}_t$ are computed as in \eqref{eq:mini_batch_grads}.
    \begin{equation}\label{eq:mini_batch_grads}
        g_i = \nabla_\theta \mathcal{L}(\theta_{t-1}, x_i),\quad x_i \in \mathcal{B}_t
    \end{equation}
    Then, the $L_2$ norm of each gradient is clipped to a maximum threshold $C$ according to \eqref{eq:DP_SGD_Clipping}.
    \begin{equation}\label{eq:DP_SGD_Clipping}
        \Bar{g}_i \gets g_i / \max\left(1, \frac{\|g_i\|_2}{C}\right) 
    \end{equation}
    This step is crucial because it limits the \textit{sensitivity} of the aggregated gradient to any individual sample. Bounding the maximum influence of each data point is a fundamental requirement for calibrating the added noise properly.
    \item \textbf{Noise Addition:} After computing the average of the clipped gradients $\Bar{g}_i$, random noise is added to the result, typically Gaussian. The noise is sampled from $\mathcal{N}(0, \sigma^2 C^2 I)$, where $I$ is the identity matrix, $C$ is the clipping threshold, and $\sigma$ is the \emph{noise multiplier}. The noise magnitude ($\propto \sigma C$) is calibrated to obscure the individual contribution of each sample in the aggregated gradient $\tilde{g}_t$, thereby providing the DP guarantee. A larger $\sigma$ implies more noise, thus stronger privacy (i.e., smaller $\varepsilon$), but potentially lower practical utility.
\end{enumerate}

Finally, the model parameters are updated using this noisy aggregated gradient as in \eqref{eq:DP_SGD_update}.
\begin{equation}\label{eq:DP_SGD_update}
    \theta_t \gets \theta_{t-1} - \eta \tilde{g}_t
\end{equation}

\begin{algorithm}[t]
\caption{Basic Differential Privacy Accounting}
\label{alg:accouting}
\begin{algorithmic}[1]
\STATE \textbf{Input:} number of steps $T$, per-step privacy parameters $(\varepsilon_t, \delta_t)$, model state $\mathcal{M}_t$
\STATE \textbf{Initialize:} $(\varepsilon_t, \delta_t) \gets (\varepsilon_0, \delta_0)$
\FOR{$t = 1$ \textbf{to} $T$}
    \STATE $\varepsilon_t \gets$ \texttt{ExtractEpsilon}$(\mathcal{M}_t)$
    \STATE $\delta_t \gets$ \texttt{ExtractDelta}$(\mathcal{M}_t)$
\ENDFOR
\STATE \textbf{Output:} accumulated privacy $\lbrace(\varepsilon_t, \delta_t)\rbrace_{t=0}^T$
\end{algorithmic}
\end{algorithm}

A fundamental aspect of iterative algorithms with DP, such as DP-SGD, is the privacy budget tracking (\textit{privacy accounting}, see Algorithm \ref{alg:accouting}). Each step of DP-SGD consumes part of the total budget $(\varepsilon, \delta)$ which must be tracked across the $T$ training iterations. Simple composition (e.g., aggregating $\varepsilon$ and $\delta$ across steps) provides valid but often overly conservative bounds. In practice, more refined techniques are used, such as Advanced Composition \cite{dwork2014algorithmic}, or more commonly, accounting based on Rényi Differential Privacy (RDP) \cite{mironov2017renyi} or Privacy Loss Distribution (PLD) analysis \cite{sommer2018privacy}, among others \cite{zhu2022optimal}. These methods, often referred to as \textit{privacy accountants}, incorporate mini-batch sampling probabilities and the noise mechanism (e.g., Gaussian) to provide much tighter bounds on the accumulated budget $(\varepsilon, \delta)$ after $T$ steps. Tools such as \href{https://opacus.ai/}{Opacus} \cite{opacus_paper} for PyTorch implement these advanced accountants as a \textit{de facto} standard.

The application of DP, and in particular DP-SGD, inevitably introduces a \textbf{privacy-utility trade-off}. Adding more noise (larger $\sigma$, smaller $\varepsilon$) to strengthen privacy guarantees, typically degrades model performance in classification or regression tasks. Finding the right balance requires careful tuning of both DP hyperparameters (e.g., $C$, $\sigma$) and training hyperparameters (e.g., $\eta$, $L$, $T$) for a target budget $(\varepsilon, \delta)$.

Additionally, \citet{lomurno2022utility} showed that some classical regularization methods (e.g., \textit{dropout} or $L_2$) behave similarly to DP-SGD in terms of privacy levels, with a substantial reduction in computational cost and impact on prediction quality. However, these techniques \textit{do not} provide the formal, quantifiable guarantees of differential privacy.

More recently, following this same line of thought, \citet{loss_function_regularization} demonstrated that the noise injection and clipping of DP-SGD can be interpreted as an implicit regularization of the loss function $\mathcal{L}$. However, it is important to note that this regularization is \textit{independent} of the model parameters $\theta$. To address this, the authors introduced a new method for differential privacy via \textbf{explicit regularization} (PDP-SGD, \textit{Proportional DP-SGD}), where the injected noise is \emph{proportional} to each parameter. In principle, this yields better privacy guarantees more efficiently, since any traditional optimization scheme can be used to train our neural network.

In the context of federated learning, differential privacy is typically applied \emph{locally}. Each client runs DP-SGD, or the chosen method, on its local data before sending updates to the server. This protects client data not only from external adversaries but also from the central server itself should it be compromised. Likewise, it also acts as a direct defense, at least \emph{in theory}, against the gradient leakage attacks discussed earlier in Section \ref{sec:GLA}, since the gradient intercepted by an attacker would have already been previously perturbed and thus should not reveal \emph{sensitive information} about the underlying training data.

\section{Methodology}

In this section, we detail the experimental methodology followed to evaluate the impact of Differential Privacy (DP) on the performance of both image classification models and on their ability to mitigate gradient leakage attacks (GLA). Our objective is twofold:

\begin{itemize}
    \item First, to compare the performance of different Computer Vision (CV) architectures trained with, and \textit{without}, differential privacy guarantees under various privacy budgets;
    \item Second, to analyze qualitatively and quantitatively how these mechanisms affect an adversary’s ability to reconstruct private training data from shared gradients, simulating a GLA attack scenario.
\end{itemize}

\subsection{Classification Using Differentially Private Models}\label{sec:methodology_dp}

To evaluate the \emph{trade-off} between model privacy and utility, we select a binary classification task with enough complexity to observe significant performance differences under different training conditions. An overly simple task, such as the MNIST dataset used for the GLA simulation, may fail to reveal performance degradations introduced by DP; while an overly complex task may hinder the training of useful models even without privacy.

\begin{figure}[t]
    \centering
    \includegraphics[width=0.8\columnwidth]{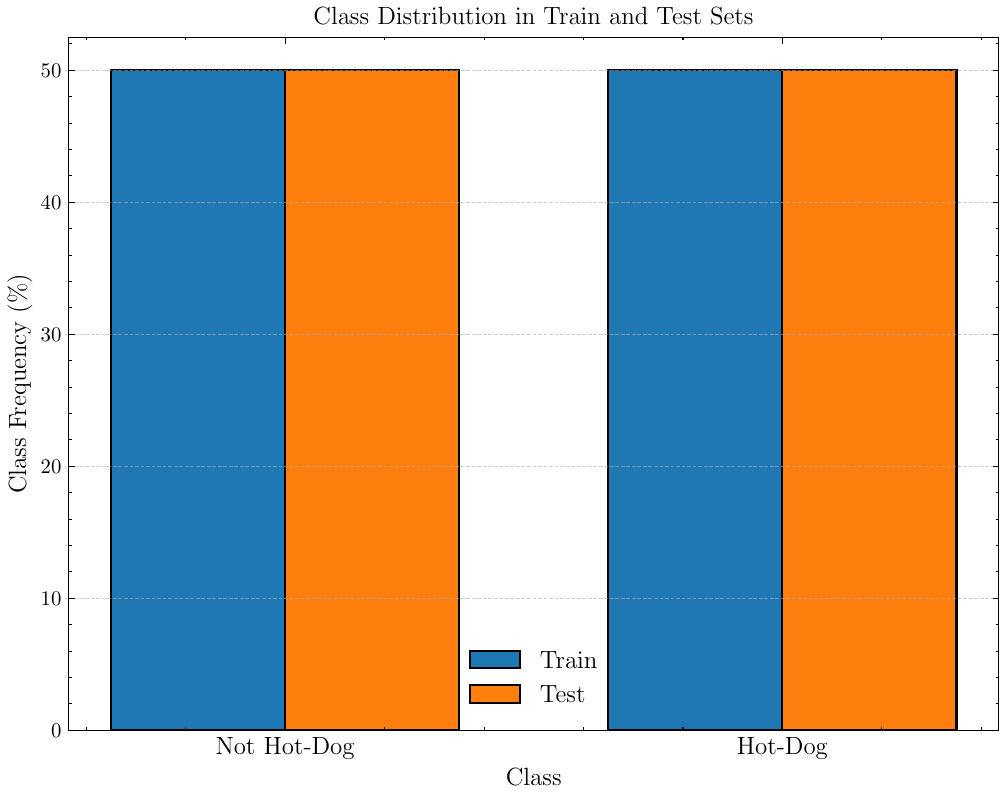}
    \caption{Class Distribution in the Training and Test Splits}
    \label{fig:class_dist}
\end{figure}

\begin{figure}[t]
    \centering
    \includegraphics[width=\columnwidth]{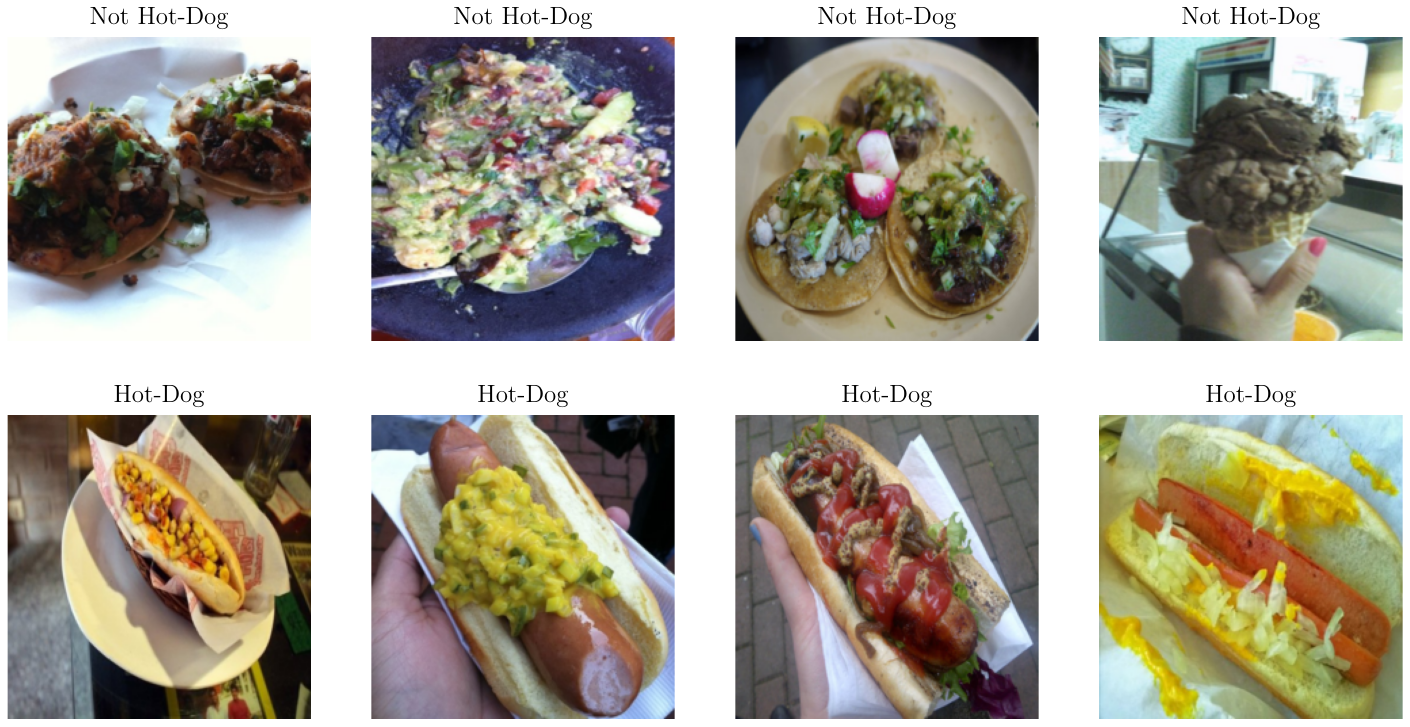}
    \caption{Examples of Training Images (\textit{Hot-Dog vs Not Hot-Dog})}
    \label{fig:class_examples}
\end{figure}

\textbf{Dataset:} We use an outcome-balanced subset of the well-known Food-101 dataset \citep{bossard2014food} focused on binary classification with an equal distribution of classes. This dataset, with representative examples shown in Figure \ref{fig:class_examples}, consists of images belonging to two classes: \textit{hot-dog} and \textit{not hot-dog}. As seen in Figure \ref{fig:class_dist}, the dataset is perfectly balanced between the two classes in both the training and evaluation sets, thus avoiding class imbalance biases.

\textbf{Compared Models:} In order to assess the impact of different differential privacy methods across model architectures and training strategies, we compare three distinct computer vision models:

\begin{itemize}
    \item \textbf{Simple Convolutional Neural Network (Custom-CNN):} A relatively small and standard CNN architecture (two convolutional layers with \textit{max-pooling}, followed by two linear layers), trained entirely from scratch. This serves as a baseline for evaluating performance without the benefit of pretraining.
    \item \textbf{ResNet50:} We employ a ResNet50 \cite{he2016deep} pretrained on ImageNet \cite{deng2009imagenet}. We apply a \textit{fine-tuning} strategy where all pretrained convolutional layers have been frozen, and only a new final classification head is trained for our task. The ResNet50 architecture is based on \emph{residual blocks}, which enable training deeper networks without suffering from vanishing gradient problems. In this case, we use the \emph{pretrained backbone} as a feature extractor and train only the final classification head. Figure \ref{fig:ResNet_arch} illustrates the general architecture of ResNet50.
    \begin{figure*}[t]
        \centering
        \includegraphics[angle=90, width=0.9\linewidth]{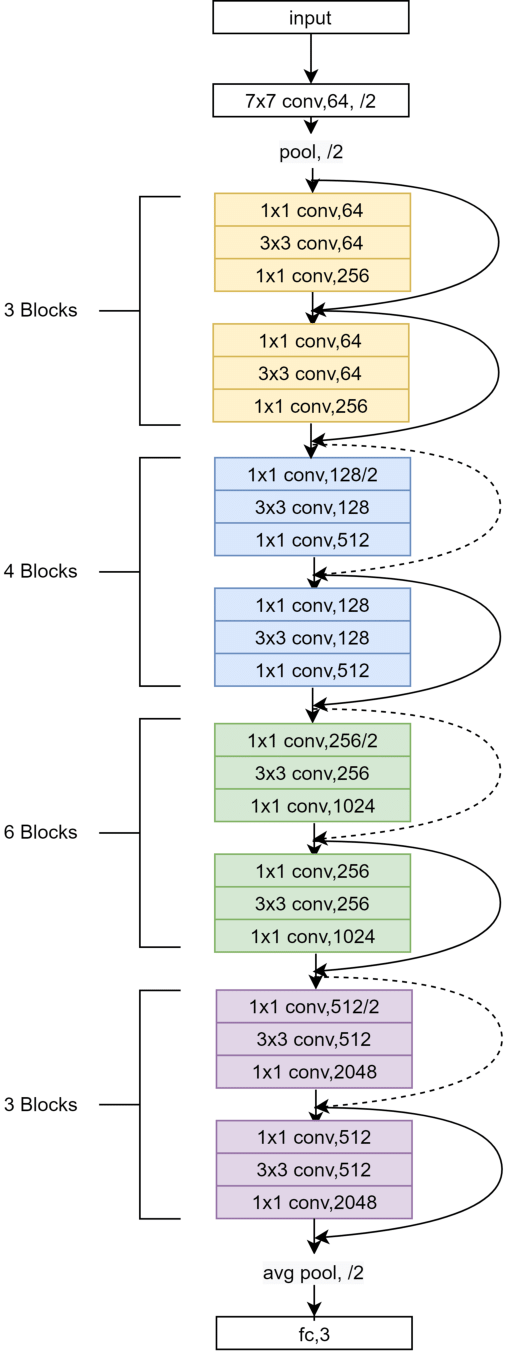}
        \caption{ResNet50 Model Architecture \cite{he2016deep}}
        \label{fig:ResNet_arch}
    \end{figure*}
    \item \textbf{Vision Transformer (ViT) DINOv2 with Registers:} Finally, we employ an advanced Vision Transformer (ViT) model, specifically DINOv2 \cite{oquab2023dinov2}, incorporating the \textit{registers} technique \cite{darcet2023vision}. Registers are additional tokens introduced during pretraining that help the model capture global scene information, improving downstream performance. More specifically, they \textit{remove artifacts} in ViT feature maps, as shown in Figure \ref{fig:vit_registers_example}, and improve classification performance. As with ResNet50, we freeze the pretrained ViT backbone and train only the final classification head for our task.
    \begin{figure*}[t]
        \centering
        \includegraphics[width=0.8\linewidth]{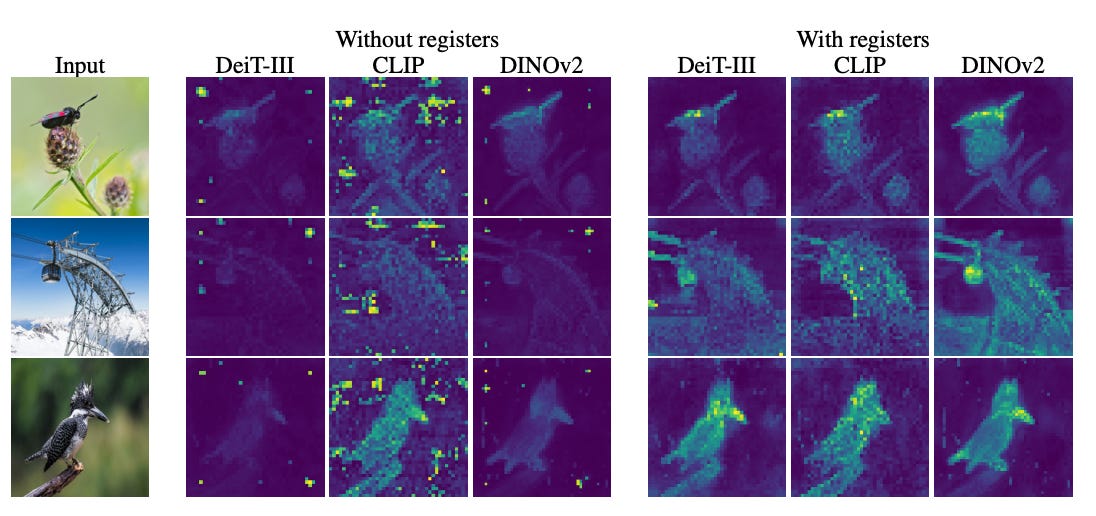}
        \caption{Example of Artifact Removal in ViT Models with Registers \cite{darcet2023vision}}
        \label{fig:vit_registers_example}
    \end{figure*}
\end{itemize}

\textbf{Training Methods:} Each of the three models described above has been trained under different regimes to compare their performance with and without differential privacy protection:

\begin{enumerate}
    \item \textbf{Standard Training (No DP):} Models have been trained using the Adam optimizer \cite{kingma2014adam} and the standard binary classification loss, \textit{Binary Cross-Entropy} (BCE) in \eqref{eq:BCE}. This serves as a baseline scenario without any formal privacy guarantees ($\varepsilon = \infty$).
    \begin{equation}\label{eq:BCE}
        \scalemath{0.9}{\mathcal{L}_{\text{BCE}} = -\frac{1}{N}\sum_{i=1}^{N}y_i\log(\hat{y}_i) + (1-y_i)\log(1-\hat{y}_i)}
    \end{equation}
    where $N$ is the total number of instances, $y_i$ is the true label, and $\hat{y}_i=f(x_i; \theta)$ is the model prediction for instance $x_i$.
    \item \textbf{DP-SGD:} We train the models using the DP-SGD algorithm, as described in Section \ref{sec:DP} (Algorithm \ref{alg:DP_SGD}). To assess the impact of different protection levels, we use three privacy budgets ($\varepsilon=8$, $\varepsilon=25$, and $\varepsilon=50$), keeping $\delta$ fixed and small ($\delta \approx 1/N_{train}$, where $N_{train}$ is the size of the training set). For the implementation, we use the \href{https://opacus.ai/}{Opacus} library by Meta \cite{opacus_paper}.
    \item \textbf{PDP-SGD (Explicit Regularization):} We also train our models using the Proportional DP-SGD (PDP-SGD) approach \cite{loss_function_regularization}, which implements DP via explicit loss function regularization, where added noise is \textbf{proportional} to model parameters. This method is included in order to compare an alternative to DP-SGD which, theoretically, should offer better results in terms of model performance. The regularized loss function proposed by PDP-SGD is given by \eqref{eq:PDP_SGD}.
    \begin{equation}\label{eq:PDP_SGD}
        \mathcal{L}_{\text{PDP}} = \mathcal{L} + \kappa \sum_{i} \theta_{i}^{2} x_{i}^{2}
    \end{equation}
    where $\mathcal{L}$ is the original loss function (e.g., binary cross-entropy in our case), and $\kappa$ is a regularization constant related to the learning rate $\eta$ and the noise parameter $\sigma$ by $\kappa=\eta^2\sigma^2$. Moreover:
    \begin{equation}
        \nabla_{\theta_{i}} \mathcal{L}_{\text{PDP}} = \nabla_{\theta_{i}} \mathcal{L} + 2\kappa x_{i}^{2} \theta_{i}
    \end{equation}
\end{enumerate}

Regarding training hyperparameters, all models have been trained with the Adam optimizer \cite{kingma2014adam}, an image size of $224\times224$ pixels (in RGB), an initial learning rate of $\eta=0.001$, and batch size $L=32$. Additionally, we use \emph{early-stopping} as a training halting criterion where training is terminated if no improvement on the validation loss is observed over 20 consecutive epochs. For DP-SGD, clipping size has been set to $C=1.2$ with a default noise multiplier. Likewise, for the explicit PDP-SGD regularization, we use a noise multiplier of $\sigma=0.1$ and a learning rate of $\eta=0.1$. In all cases, the validation was carried out on a set comprising 20\% of the total training data.

\subsection{Reconstruction Attacks via Gradient Leakage}\label{sec:methodology_gla}

The second experimental component of this work focuses on the simulation and analysis of reconstruction attacks based on gradient leakages. Our goal is to investigate the feasibility and limitations of recovering private training data --- specifically, images used to train a computer vision model, for example --- from gradients intercepted in a federated learning environment under the influence of differential privacy methods.

\textbf{Dataset \& Model:} For this stage, we use the \textbf{MNIST} dataset \cite{lecun1998gradient}, a standard benchmark for image classification tasks consisting of grayscale handwritten digits. In this case, the images have been resized to $32 \times 32$ pixels, and the victim model architecture is a simple Convolutional Neural Network (CNN), composed of four convolutional layers followed by a linear (dense) layer for classification. This architecture, while simple, is sufficient to demonstrate the feasibility of gradient leakage attacks in a controlled environment.

\textbf{Reconstruction Process:} The attack aims to reconstruct an image $x_{\text{rec}}$ and its corresponding label $y_{\text{rec}}$ such that the gradients generated by this pair $(\nabla \theta_{\text{att}})$ match the original intercepted gradients $(\nabla \theta_t^{(k)})$ as closely as possible, as described in Equation \ref{eq:gla_objective}. In our case, we \textit{simultaneously optimize} both the reconstructed image and the label by employing an L-BFGS second-order optimizer \cite{liu1989limited}. As mentioned in Section \ref{sec:GLA}, the label can also be inferred directly from the intercepted gradients, but in our experiments, we found that jointly optimizing both the image and label yields better reconstruction results.

\textbf{Victim Model Training Scenarios:} To evaluate the impact of differential privacy methods as a safeguard against GLAs, the reconstruction attack is performed on gradients obtained from the victim CNN, trained (more on this later) under three different conditions, analogous to those used in the classification task (Section \ref{sec:methodology_dp}):
\begin{enumerate}
    \item \textbf{Standard Training:} Gradients from a model trained in a standard manner, without any differential privacy protection, are used. This serves as a baseline to determine the maximum possible information leakage in this scenario.
    \item \textbf{DP-SGD Training:} Gradients are generated from a model trained using DP-SGD. The same privacy budgets (with $\delta \approx 1/N_{train}$) and DP hyperparameters as in the classification experiments are explored.
    \item \textbf{PDP-SGD Training (Explicit Regularization):} Gradients are obtained from a model trained with the Proportional DP-SGD approach. The same regularization hyperparameters (e.g., $\kappa$ derived from $\eta=0.1$ and $\sigma=0.1$) used in the classification task are employed.
\end{enumerate}

\textbf{Considerations on Attack Stability:} One of the main challenges of gradient leakage reconstruction attacks is their \emph{instability}. The reconstruction optimization process is highly sensitive to initialization and may converge to suboptimal critical points (local minima or saddle points) which do not correspond to the original data, or even \emph{diverge} completely, especially if the initialization is inadequate or the search space is too complex. This lack of reliability is a significant limitation and was a determining factor in choosing a relatively simple dataset and model architecture (MNIST, small CNN), with small, black-and-white images.

To partially mitigate these stability and convergence issues, we initialize the neural network with a weight configuration that leads to a \emph{stable reconstruction}. In practice, this is equivalent to assuming that the network has already been trained for a few epochs and that its weights are in a region of the parameter space that is conducive to convergence. This assumption allows us to focus on the information leakage mechanism through gradients without the results being obscured by an early divergence of the attack's optimizer due to poor model initialization.

Finally, although the general context of this work is framed within federated learning, for this specific GLA simulation, we assume that the gradients have been \emph{intercepted} in a federated learning environment, but in practice, the attack is performed on a single, centrally trained model. This allows us to study the information leakage process from an individual client's gradient without the added complexity of multi-client aggregation or federated communication protocols. The goal is to demonstrate the potential for information leakage at the level of a single gradient, which is the fundamental unit of information shared in many FL schemes.

\section{Results \& Discussion}

In this section, we present and analyze the results derived from the experiments described in Sections \ref{sec:methodology_dp} and \ref{sec:methodology_gla}. The content is organized as follows: first, we describe the \emph{evaluation metrics} used to quantify both model performance and the quality of reconstructions (Section \ref{ssec:evaluation_metrics}). Next, we report the results obtained in the \emph{classification task} under several differential privacy regimes, comparing the utility of various privacy levels and protection mechanisms on the performance of vision models (Section \ref{ssec:classification_results_dp}). Finally, we present the results of the reconstruction attacks via \emph{gradient leakage}, where we assess both qualitatively and quantitatively the quality of the reconstructed images under the influence of DP-SGD and PDP-SGD, and discuss their effectiveness as defense strategies against the recovery of sensitive information (Section \ref{ssec:results_gla}).

All experiments were run using PyTorch and Python~3.12. Given the modest scale of the models and datasets, the experiments can also be reproduced on any standard consumer-grade GPUs.

\subsection{Evaluation Metrics}\label{ssec:evaluation_metrics}

To quantitatively evaluate the results obtained in the two main experimental tasks of this work --- image classification and gradient-based reconstruction attacks --- we employ different sets of metrics adapted to each objective.

For the \textbf{binary classification task} (\textit{Hot-Dog vs Not Hot-Dog}), we use the standard metrics commonly applied to classification problems, as presented in Table \ref{tab:results_classification}. These metrics are derived from the entries of the confusion matrix: True Positives (TP), True Negatives (TN), False Positives (FP), and False Negatives (FN). Specifically, these metrics are:

\begin{itemize}
    \item \textbf{Accuracy:} The overall proportion of correct predictions.
    \begin{equation}
        \text{Accuracy} = \dfrac{\text{TP} + \text{TN}}{\text{TP} + \text{TN} + \text{FP} + \text{FN}}
    \end{equation}
    \item \textbf{Precision:} Among all instances predicted as positive, how many were truly positive?
    \begin{equation}
        \text{Precision} = \dfrac{\text{TP}}{\text{TP} + \text{FP}}
    \end{equation}
    \item \textbf{Recall (Sensitivity):} Of all instances that were truly positive, how many were correctly identified?
    \begin{equation}
        \text{Recall} = \dfrac{\text{TP}}{\text{TP} + \text{FN}}
    \end{equation}
    \item \textbf{Specificity:} Of all instances that were truly negative, how many were correctly identified?
    \begin{equation}
        \text{Specificity} = \dfrac{\text{TN}}{\text{TN} + \text{FP}}
    \end{equation}
    \item \textbf{$F_1$-Score:} The harmonic mean of Precision and Recall, providing a balance between the two. This is especially useful in cases of class imbalance (although that is not the case in this dataset).
    \begin{equation}
        F_1 = 2 \cdot \dfrac{\text{Precision} \cdot \text{Recall}}{\text{Precision} + \text{Recall}}
    \end{equation}
    \item \textbf{MCC (Matthews Correlation Coefficient):} A correlation coefficient between the observed and predicted classifications. Considered a robust and balanced metric even with imbalanced classes. It ranges from $-1$ (total disagreement) to $+1$ (perfect agreement), with 0 indicating random correlation.
    \begin{equation}
        \scalemath{0.8}{\text{MCC} = \dfrac{\text{TP} \cdot \text{TN} - \text{FP} \cdot \text{FN}}{\sqrt{(\text{TP}+\text{FP})(\text{TP}+\text{FN})(\text{TN}+\text{FP})(\text{TN}+\text{FN})}}}
    \end{equation}
\end{itemize}

For the task of \textbf{image reconstruction} from the gradient leakage (GLA), the goal is to evaluate the similarity between the original image $x$ and the reconstructed image $x_{\text{rec}}$. While metrics such as the Mean Squared Error (MSE) are common, they do not always reflect \textit{human-perceived similarity}. Therefore, in this work, we focus on the \textbf{Structural Similarity Index Measure (SSIM)} \cite{wang2004image}, specifically designed to capture structural similarity between two images in a way that is more \textit{consistent} with human visual perception.

SSIM evaluates similarity based on three comparative components between the two images ($x$ and $x'$), typically computed over local windows and then averaged:

\begin{enumerate}
    \item \textbf{Luminance:} Compares the mean brightness of the images, based on pixel intensity means ($\mu_x, \mu_{x'}$):
    \begin{equation}
        l(x, x') = \dfrac{2\mu_x \mu_{x'} + c_1}{\mu_x^2 + \mu_{x'}^2 + c_1}
    \end{equation}
    \item \textbf{Contrast:} Compares the variation in intensity (dynamic range) around the mean, based on the standard deviations ($\sigma_x, \sigma_{x'}$):
    \begin{equation}
        c(x, x') = \dfrac{2\sigma_x \sigma_{x'} + c_2}{\sigma_x^2 + \sigma_{x'}^2 + c_2}
    \end{equation}
    \item \textbf{Structure:} Compares the structural correlation between the images, based on covariance ($\sigma_{xx'}$):
    \begin{equation}
        s(x, x') = \dfrac{\sigma_{xx'} + c_3}{\sigma_x \sigma_{x'} + c_3}
    \end{equation}
\end{enumerate}
where $c_1 = (k_1 L)^2$, $c_2 = (k_2 L)^2$, and $c_3 = c_2/2$ are stabilization constants to avoid division by zero, with $L$ being the dynamic range of the pixel values (e.g., 255 for 8-bit images), and $k_1, k_2$ are two small constants (typically $k_1=0.01, k_2=0.03$).

The global SSIM index is then defined as a combination of these three components:
\begin{equation}
    \text{SSIM}(x, x') = [l(x, x')]^\alpha \cdot [c(x, x')]^\beta \cdot [s(x, x')]^\gamma
\end{equation}
where $\alpha > 0, \beta > 0, \gamma > 0$ are parameters controlling the relative importance of each component. Commonly, $\alpha=\beta=\gamma=1$, simplifying the formula to:
\begin{equation}
    \text{SSIM}(x, x') = \frac{(2\mu_x \mu_{x'} + c_1)(2\sigma_{xx'} + c_2)}{(\mu_x^2 + \mu_{x'}^2 + c_1)(\sigma_x^2 + \sigma_{x'}^2 + c_2)}
\end{equation}
The SSIM value ranges between $-1$ and $1$, where $1$ indicates perfect structural similarity between the two images. Higher SSIM values indicate visually higher-quality reconstructions in the context of GLA, meaning that the attacker has recovered an image more closely resembling the original.

\subsection{Classification with Differentially Private Models}\label{ssec:classification_results_dp}

\begin{table*}[t]
  \centering\caption{Classification Results Across 10 Independent Executions (Values Reported as Mean $\pm$ Standard Error)}
  \renewcommand{\arraystretch}{1.5}
  \begin{tabular}{lcccccccc}
  \toprule
   & \textbf{$\varepsilon$} & \textbf{Accuracy} & \textbf{Precision} & \textbf{Recall} & \textbf{Specificity} & \textbf{$F_1$} & \textbf{MCC} \\
  \midrule
  \multicolumn{8}{c}{\textbf{Convolutional Neural Network (Custom-CNN)}} \\
  \midrule
  \textbf{Standard Training} & $\infty$ & 0.584\textsubscript{\scriptsize $\pm$0.005} & 0.586\textsubscript{\scriptsize $\pm$0.006} & 0.572\textsubscript{\scriptsize $\pm$0.010} & 0.596\textsubscript{\scriptsize $\pm$0.011} & 0.579\textsubscript{\scriptsize $\pm$0.006} & 0.168\textsubscript{\scriptsize $\pm$0.009} \\
  \textbf{DP-SGD $(\varepsilon=8)$} & 8 & 0.562\textsubscript{\scriptsize $\pm$0.013} & 0.629\textsubscript{\scriptsize $\pm$0.030} & 0.342\textsubscript{\scriptsize $\pm$0.151} & 0.682\textsubscript{\scriptsize $\pm$0.152} & 0.428\textsubscript{\scriptsize $\pm$0.088} & 0.131\textsubscript{\scriptsize $\pm$0.015} \\
  \textbf{DP-SGD $(\varepsilon=25)$} & 25 &  0.598\textsubscript{\scriptsize $\pm$0.015} & 0.612\textsubscript{\scriptsize $\pm$0.024} & 0.536\textsubscript{\scriptsize $\pm$0.118} & 0.660\textsubscript{\scriptsize $\pm$0.153} & 0.571\textsubscript{\scriptsize $\pm$0.038} & 0.198\textsubscript{\scriptsize $\pm$0.018} \\
  \textbf{DP-SGD $(\varepsilon=50)$} & 50 & 0.588\textsubscript{\scriptsize $\pm$0.010} & 0.655\textsubscript{\scriptsize $\pm$0.019} & 0.372\textsubscript{\scriptsize $\pm$0.019} & 0.804\textsubscript{\scriptsize $\pm$0.033} & 0.475\textsubscript{\scriptsize $\pm$0.009} & 0.195\textsubscript{\scriptsize $\pm$0.022} \\
  \textbf{Explicit Regularization} & -- & 0.614\textsubscript{\scriptsize $\pm$0.013} & 0.609\textsubscript{\scriptsize $\pm$0.019} & 0.636\textsubscript{\scriptsize $\pm$0.011} & 0.592\textsubscript{\scriptsize $\pm$0.030} & 0.622\textsubscript{\scriptsize $\pm$0.009} & 0.228\textsubscript{\scriptsize $\pm$0.028} \\
  \midrule
  \multicolumn{8}{c}{\textbf{Pretrained ResNet50 (\textit{Fine-Tuning} of Classification Head Only)}} \\
  \midrule
  \textbf{Standard Training} & $\infty$ & 0.886\textsubscript{\scriptsize $\pm$0.003} & 0.873\textsubscript{\scriptsize $\pm$0.005} & 0.904\textsubscript{\scriptsize $\pm$0.002} & 0.868\textsubscript{\scriptsize $\pm$0.006} & 0.888\textsubscript{\scriptsize $\pm$0.003} & 0.773\textsubscript{\scriptsize $\pm$0.007} \\
  \textbf{DP-SGD $(\varepsilon=8)$} & 8 & 0.844\textsubscript{\scriptsize $\pm$0.006} & 0.831\textsubscript{\scriptsize $\pm$0.015} & 0.864\textsubscript{\scriptsize $\pm$0.018} & 0.824\textsubscript{\scriptsize $\pm$0.020} & 0.847\textsubscript{\scriptsize $\pm$0.006} & 0.689\textsubscript{\scriptsize $\pm$0.012} \\
  \textbf{DP-SGD $(\varepsilon=25)$} & 25 & 0.868\textsubscript{\scriptsize $\pm$0.005} & 0.871\textsubscript{\scriptsize $\pm$0.008} & 0.864\textsubscript{\scriptsize $\pm$0.012} & 0.872\textsubscript{\scriptsize $\pm$0.009} & 0.868\textsubscript{\scriptsize $\pm$0.005} & 0.736\textsubscript{\scriptsize $\pm$0.009} \\
  \textbf{DP-SGD $(\varepsilon=50)$} & 50 & 0.866\textsubscript{\scriptsize $\pm$0.003} & 0.877\textsubscript{\scriptsize $\pm$0.011} & 0.852\textsubscript{\scriptsize $\pm$0.015} & 0.880\textsubscript{\scriptsize $\pm$0.011} & 0.864\textsubscript{\scriptsize $\pm$0.004} & 0.732\textsubscript{\scriptsize $\pm$0.004} \\
  \textbf{Explicit Regularization} & -- & 0.892\textsubscript{\scriptsize $\pm$0.002} & 0.883\textsubscript{\scriptsize $\pm$0.004} & 0.904\textsubscript{\scriptsize $\pm$0.003} & 0.880\textsubscript{\scriptsize $\pm$0.005} & 0.893\textsubscript{\scriptsize $\pm$0.002} & 0.784\textsubscript{\scriptsize $\pm$0.004} \\
  \midrule
  \multicolumn{8}{c}{\textbf{Pretrained DINOv2 w/ Registers (Vision Transformer, \textit{Fine-Tuning} of Classification Head Only)}} \\
  \midrule
  \textbf{Standard Training} & $\infty$ & 0.972\textsubscript{\scriptsize $\pm$0.004} & 0.992\textsubscript{\scriptsize $\pm$0.002} & 0.952\textsubscript{\scriptsize $\pm$0.007} & 0.992\textsubscript{\scriptsize $\pm$0.001} & 0.971\textsubscript{\scriptsize $\pm$0.004} & 0.945\textsubscript{\scriptsize $\pm$0.007} \\
  \textbf{DP-SGD $(\varepsilon=8)$} & 8 & 0.964\textsubscript{\scriptsize $\pm$0.002} & 0.992\textsubscript{\scriptsize $\pm$0.002} & 0.936\textsubscript{\scriptsize $\pm$0.003} & 0.992\textsubscript{\scriptsize $\pm$0.001} & 0.963\textsubscript{\scriptsize $\pm$0.002} & 0.929\textsubscript{\scriptsize $\pm$0.003} \\
  \textbf{DP-SGD $(\varepsilon=25)$} & 25 & 0.968\textsubscript{\scriptsize $\pm$0.003} & 0.996\textsubscript{\scriptsize $\pm$0.001} & 0.940\textsubscript{\scriptsize $\pm$0.007} & 0.996\textsubscript{\scriptsize $\pm$0.001} & 0.967\textsubscript{\scriptsize $\pm$0.004} & 0.938\textsubscript{\scriptsize $\pm$0.007} \\
  \textbf{DP-SGD $(\varepsilon=50)$} & 50 & 0.968\textsubscript{\scriptsize $\pm$0.001} & 0.996\textsubscript{\scriptsize $\pm$0.001} & 0.942\textsubscript{\scriptsize $\pm$0.004} & 0.996\textsubscript{\scriptsize $\pm$0.001} & 0.968\textsubscript{\scriptsize $\pm$0.001} & 0.939\textsubscript{\scriptsize $\pm$0.003} \\
  \textbf{Explicit Regularization} & -- & 0.976\textsubscript{\scriptsize $\pm$0.003} & 0.990\textsubscript{\scriptsize $\pm$0.002} & 0.962\textsubscript{\scriptsize $\pm$0.005} & 0.990\textsubscript{\scriptsize $\pm$0.002} & 0.976\textsubscript{\scriptsize $\pm$0.003} & 0.953\textsubscript{\scriptsize $\pm$0.005} \\
  \bottomrule
  \end{tabular}
  \label{tab:results_classification}
\end{table*}

The quantitative results of the \textbf{image classification} task (\textit{Hot-Dog vs Not Hot-Dog}), using the different architectures and training methods with and without differential privacy, and evaluated with the metrics presented in Section \ref{ssec:evaluation_metrics}, are summarized in Table \ref{tab:results_classification}. The analysis of these results reveals some general trends regarding the \textit{trade-off} between the strength of privacy guarantees and model utility (or performance), as well as some noteworthy differences between DP-SGD and explicit regularization (PDP-SGD).

Overall, and as expected, the introduction of differential privacy through \textbf{DP-SGD tends to degrade model performance} compared to the traditionally trained alternatives (without DP, $\varepsilon=\infty$). This degradation manifests itself in most evaluation metrics (Accuracy, $F_1$-Score, MCC) and is generally more pronounced the stricter the privacy budget (i.e., the smaller the value of $\varepsilon$). This phenomenon is a direct consequence of the two core mechanisms behind DP-SGD: gradient clipping, which limits the magnitude of individual updates; and Gaussian noise injection, which obscures the contribution of each data point, potentially interfering with optimal model learning.

For the Simple Convolutional Neural Network (Custom-CNN) trained from scratch, the standard training established a modest baseline (e.g., Accuracy $0.584$, MCC $0.168$). Applying DP-SGD with $\varepsilon=8$ results in a significant performance drop (Accuracy $0.562$, MCC $0.131$), particularly driven by an extremely low Recall ($0.342$), suggesting that the model becomes overly conservative in positive predictions under strong privacy constraints. However, we observe a \emph{notable exception} with this architecture: for $\varepsilon=25$ and $\varepsilon=50$, performance (in terms of Accuracy and MCC) is slightly higher or comparable to that of standard training (e.g., Accuracy $0.598$, MCC $0.198$ for $\varepsilon=25$). This anomaly may be attributable to the simplicity of the network and the nature of the dataset; it is possible that the noise and clipping inherent to DP-SGD act as an unintended form of regularization. On the other hand, \textbf{explicit regularization (PDP-SGD) consistently improves all metrics} for the Custom-CNN (Accuracy $0.614$, $F_1$-Score $0.622$, MCC $0.228$), outperforming both the baseline and all DP-SGD variants. This suggests that its mechanism, which in practice resembles an adaptive $L_2$ regularization, is beneficial in helping the model generalize better.

In the case of ResNet50 (pretrained with fine-tuning of the classification head), the standard model already achieves considerably high performance (Accuracy $0.886$, MCC $0.773$), owing to the pretrained features learned from ImageNet. Here, the introduction of DP-SGD also follows the general trend of degrading performance. For example, with $\varepsilon=8$, Accuracy falls to $0.844$ and MCC to $0.689$. As $\varepsilon$ increases (\textit{weaker privacy}), performance progressively recovers, approaching but not fully reaching the baseline (e.g., Accuracy $0.866$ with $\varepsilon=50$). This behavior is expected, as the added noise has a more noticeable effect when the model is already operating at high accuracy. Similarly to the Custom-CNN, explicit regularization (PDP-SGD) achieves a slight improvement over the baseline (Accuracy $0.892$, MCC $0.784$).

Finally, for the DINOv2 with Registers (pretrained Vision Transformer), the \textit{most advanced} architecture, the standard baseline delivers exceptional performance (Accuracy $0.972$, MCC $0.945$). The application of DP-SGD, as in the case of ResNet50, consistently reduces performance, although the absolute magnitude of the drop is smaller in percentage terms given the very high starting point (e.g., Accuracy $0.964$ with $\varepsilon=8$). Improvement is observed when moving from $\varepsilon=8$ to $\varepsilon=25$, but the results for $\varepsilon=25$ and $\varepsilon=50$ are nearly identical (Accuracy $0.968$ in both cases, MCC $0.938$ and $0.939$ respectively), suggesting that, for this model and with the DP hyperparameters used, a \emph{privacy-utility plateau} is reached around these privacy budgets. Interestingly, explicit regularization (PDP-SGD) achieves a \textit{slight improvement} over standard training (Accuracy $0.976$ vs $0.972$, MCC $0.953$ vs $0.945$), demonstrating that parameter-proportional regularization can still provide benefits even for such advanced pretrained models. In fact, the additional $L_2$-type regularization introduced by PDP-SGD appears to consistently aid generalization across all architectures, likely by preventing overfitting and promoting smoother decision boundaries. However, its relative benefit diminishes as the base model's performance increases (e.g., DINOv2), indicating that highly capable architectures may already possess sufficient implicit regularization through their design and pretraining.

\begin{figure*}[!t]
    \centering
    \includegraphics[width=\linewidth]{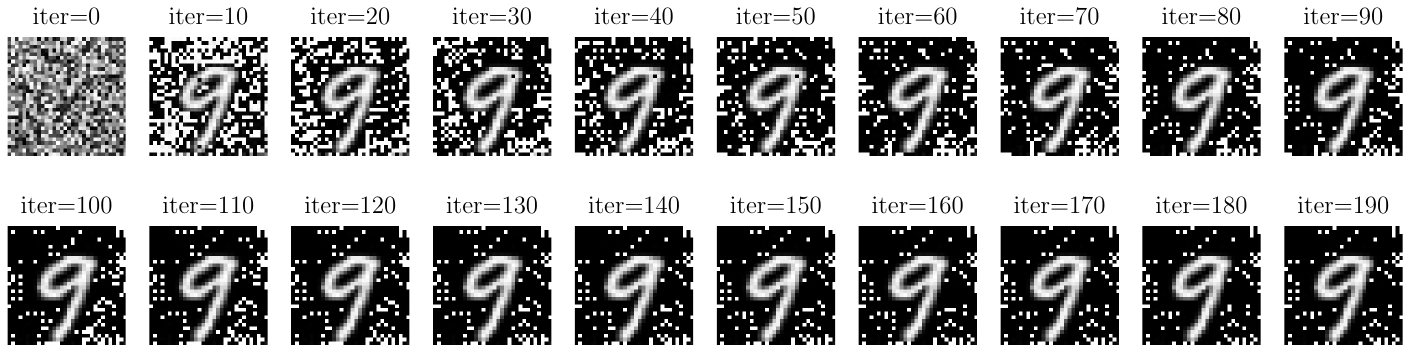}
    \caption{Reconstruction Progression of a Training Example from Model Gradients (\emph{No Privacy})}
    \label{fig:reconstructed_images}
\end{figure*}

In summary, the classification results presented in Table \ref{tab:results_classification} confirm that DP-SGD generally leads to a depreciation in model utility. The exception observed in the Custom-CNN for certain $\varepsilon$ values highlights that the interaction between DP-SGD, the architecture of the model, and complexity of the task at hand can be non-trivial. Conversely, explicit regularization (PDP-SGD) emerges as a promising alternative, improving or at least maintaining performance in most cases, likely because its effect is analogous to well-calibrated $L_2$ regularization that aids \textit{generalization}. However, its benefit may be null or even slightly negative in models that are already highly effective and robust due to their architecture and pretraining. We now turn to how these classification results translate into the effectiveness of reconstruction attacks via gradient leakage (GLA) in the next section.

\subsection{Reconstruction Attacks via Gradient Leakage}\label{ssec:results_gla}

In this section, we delve into the results of the reconstruction attack via gradient leakage (GLA), evaluating the ability of an adversary to recover images from the MNIST dataset using gradients obtained from a victim CNN. We analyze how the different training regimes --- traditional (no privacy), DP-SGD, and PDP-SGD --- impact the quality of such reconstructions. As mentioned in Section \ref{sec:methodology_gla}, it is crucial to recall that these experiments start from a neural network initialization that, in the baseline case without privacy, leads to a \emph{stable reconstruction trajectory}. In this way, we aim to evaluate the effectiveness of the reconstruction attacks without any optimization instability interfering with the results.

\begin{figure}[t]
    \centering
    \includegraphics[width=0.32\columnwidth]{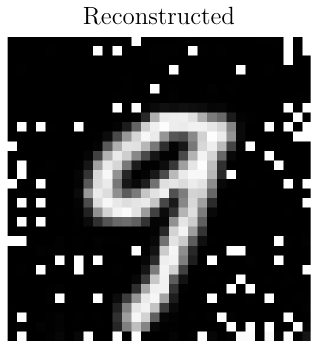}
    \includegraphics[width=0.32\columnwidth]{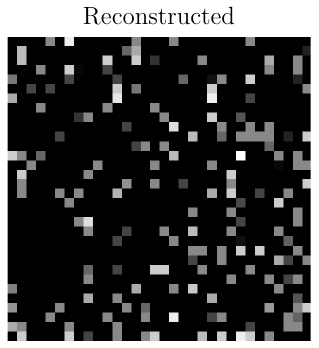}
    \includegraphics[width=0.32\columnwidth]{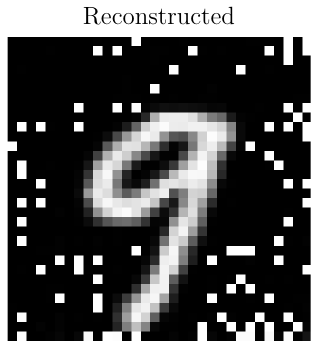}
    \caption{Final Reconstruction of a Training Instance from the Leaked Model Gradients (\textit{From Left to Right:} No Privacy vs DP-SGD vs PDP-SGD)}
    \label{fig:final_reconstructed_images}
\end{figure}

First, Figure \ref{fig:reconstructed_images} illustrates the progression of the \emph{reconstruction process} over 200 iterations for a sample image when gradients from a traditionally trained model (without any privacy protection) are used. Starting from an image of random noise, the attack optimizer (L-BFGS in our case) progressively refines the \textit{seed image} until obtaining a reconstruction that visually resembles the original. The final reconstruction quality in this \textit{privacy-free} scenario can be observed in the leftmost image of Figure \ref{fig:final_reconstructed_images}. While the reconstruction is not identical to the original, the attack successfully recovers a visually similar image with recognizable details and features, albeit containing clearly visible artifacts and noise.

\begin{figure}[t]
    \centering
    \includegraphics[width=0.9\columnwidth]{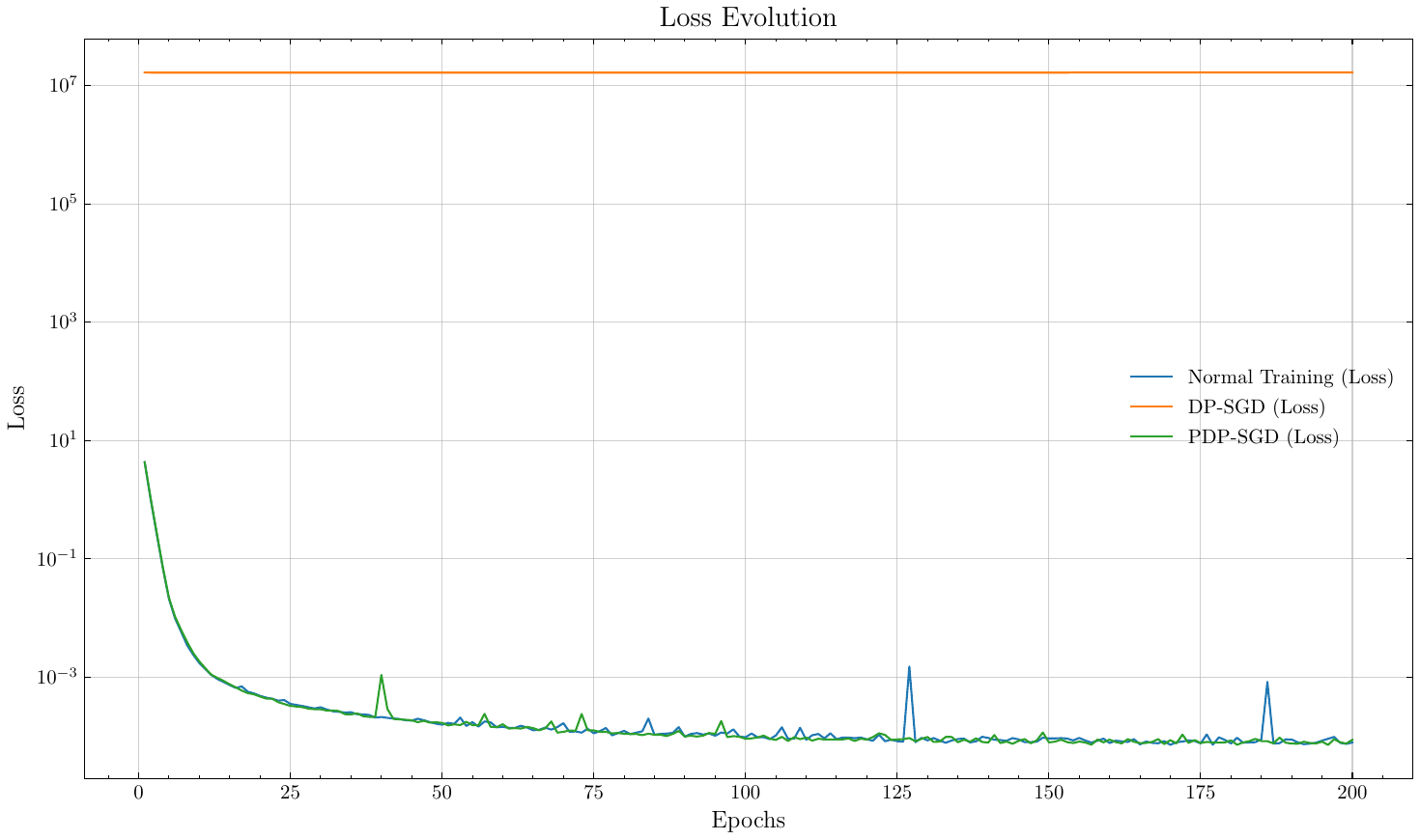}
    \caption{Evolution of the \textit{Gradient Reconstruction Loss} Across Attack Iterations}
    \label{fig:loss_evolution}
\end{figure}

\begin{figure}[t]
    \centering
    \includegraphics[width=0.9\columnwidth]{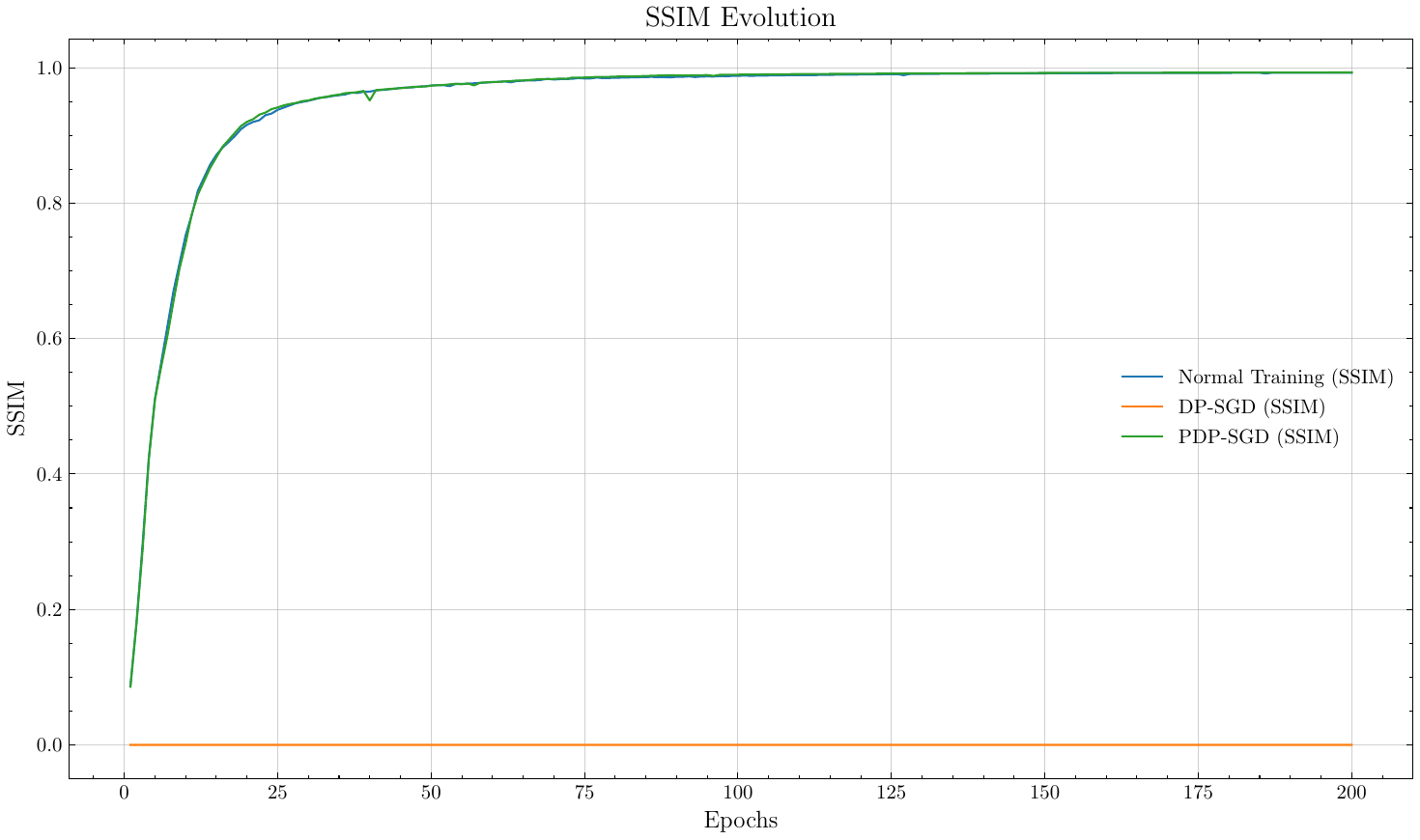}
    \includegraphics[width=0.9\columnwidth]{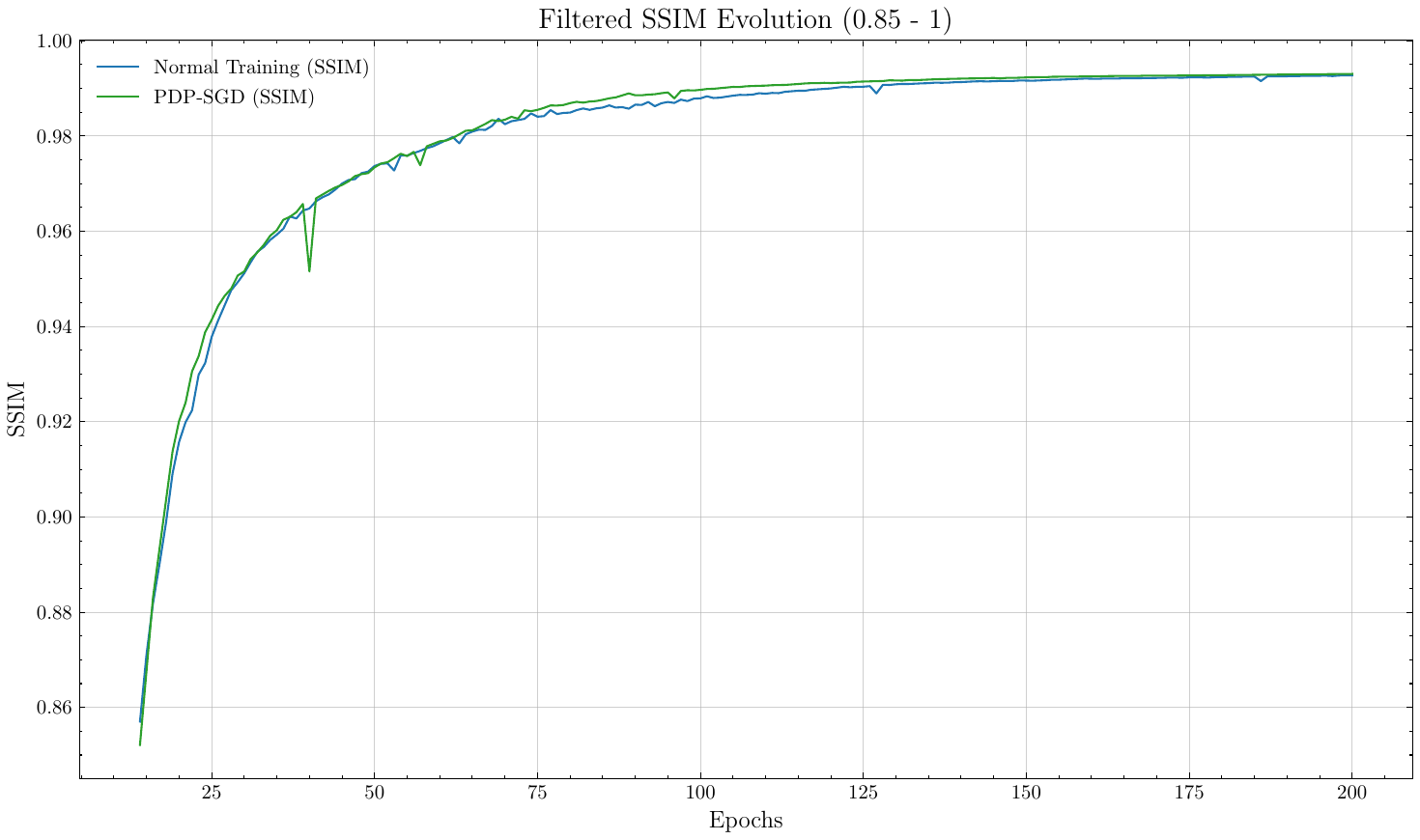}
    \caption{Evolution of the \textit{Structural Similarity Index} (SSIM) Across Attack Iterations (\textit{Note:} The lower plot zooms in on the [0.85, 1] range to better visualize convergence in the region)}
    \label{fig:SSIM_evolution}
\end{figure}

The attack optimization behavior is quantitatively reflected in Figures \ref{fig:loss_evolution} and \ref{fig:SSIM_evolution}. For traditional training (blue curve), the \emph{gradient reconstruction loss} in Figure \ref{fig:loss_evolution} decreases exponentially towards values close to zero, indicating that the attack optimizer successfully aligns the gradients of the seed image with the intercepted ones. At the same time, the SSIM index (Figure \ref{fig:SSIM_evolution}, blue curve) rises rapidly, reaching values near 1, confirming a high structural similarity between the original and reconstructed images. This clearly demonstrates the vulnerability of sharing unprotected gradients, as any potentially malicious attacker can recover sensitive information from training data.

When introducing DP-SGD during the victim model’s \textit{training} (using the same privacy budgets and hyperparameters as in the classification section, though results shown correspond to a representative $\varepsilon$, e.g., $\varepsilon=8$, which already provides strong protection), the situation changes radically. The central image in Figure \ref{fig:final_reconstructed_images} displays the result of an attempted reconstruction under DP-SGD. It is evident that \emph{the attack fails} to recover any meaningful information about the original image; the output is essentially indistinguishable noise. This failure is corroborated quantitatively: the reconstruction loss for DP-SGD (orange curve in Figure \ref{fig:loss_evolution}) stagnates at a much higher value, while the SSIM (orange curve in Figure \ref{fig:SSIM_evolution}) remains very low, indicating no structural similarity. This is an expected and desirable outcome, as it demonstrates the effectiveness of DP-SGD in perturbing gradients to obscure sensitive information over individual inputs, thereby preventing their reconstruction.

The most \emph{surprising} and noteworthy finding arises when analyzing the results of attacks on gradients protected with PDP-SGD. Contrary to what might be expected from a mechanism designed to provide differential privacy, the image reconstructed under PDP-SGD (rightmost image in Figure \ref{fig:final_reconstructed_images}) is \emph{practically identical} to that obtained in the no-privacy scenario. The reconstruction loss and SSIM curves for PDP-SGD (green curves in Figures \ref{fig:loss_evolution} and \ref{fig:SSIM_evolution}) follow trajectories almost indistinguishable from those of traditional training. The attack minimizes gradient distance and maximizes structural similarity just as effectively as if no protection were in place. This result is particularly curious when contrasted with the claims of the original work introducing PDP-SGD \cite{loss_function_regularization}, which suggests that PDP-SGD, acting as explicit parameter-proportional regularization, not only offers differential privacy guarantees but may do so more efficiently than DP-SGD. 

However, our gradient reconstruction experiments indicate that, at least under the tested configuration and in the context of image reconstruction, PDP-SGD does not provide \textit{any discernible protection} against this type of attack, behaving very similarly to standard training. A possible explanation, though requiring deeper investigation, may lie in the nature of the regularization imposed by PDP-SGD. While it formally satisfies the definition of DP, the way it perturbs (or fails to sufficiently perturb) the information seems fundamentally different from the clipping and noise addition of DP-SGD. Because PDP-SGD injects noise proportional to parameters, for small parameters the effective perturbation may be minimal, or the noise structure may fail to obscure the crucial features exploited by the attack.

In short, image reconstruction results from gradients reveal a significant vulnerability in traditional training, where an attacker can recover sensitive data. DP-SGD, on the other hand, proves highly effective in \textit{mitigating} this threat, at least in the context of image reconstruction; whereas PDP-SGD, while promising in terms of regularization and differential privacy, does not appear to provide effective protection against gradient reconstruction attacks in its current form. This raises serious questions about the applicability and effectiveness of PDP-SGD as a robust privacy-preserving mechanism in practical scenarios. 

Finally, beyond the effectiveness of the tested attacks, the \emph{sensitivity} of these methods to getting stuck in local optima or diverging during attack optimization is noteworthy. While the literature documents GLA reconstruction attacks as robust and convergent; in our experiments, we frequently observe optimization trajectories diverging or stalling in local optima, despite using a second-order optimizer such as L-BFGS. Moreover, in multiple additional tests not documented in this paper, we consistently found that these attacks face serious limitations with relatively large or multi-channel images. In fact, \citet{zhu2019deep} had already documented the \emph{limitations} of these attacks for images larger than $64 \times 64$ pixels. In any case, we believe this does not diminish the significance of our results; rather, it highlights the need for further investigation into the effectiveness of gradient leakage reconstruction and other adversarial attacks in the context of differential privacy.

\section{Conclusion}

In this work, we have focused on the critical intersection between Federated Learning (FL), Gradient Leakage Attacks (GLA), and Differential Privacy (DP), with the main objective of evaluating the effectiveness of DP mechanisms, in particular DP-SGD and an alternative based on explicit regularization (PDP-SGD), as defenses against GLAs, as well as their impact on the utility of ML models. 

The results obtained in the image classification task broadly confirm a well-known \emph{privacy-utility trade-off} when employing DP-SGD. As stricter privacy constraints are imposed, model performance tends to degrade; although exceptions have been observed with simpler architectures, where DP-SGD may act as a beneficial form of regularization under certain privacy budgets. On the other hand, PDP-SGD emerges as an interesting alternative, improving or maintaining classifier performance in all cases.

With respect to the simulation of gradient leakage attacks on the MNIST dataset, the findings are somewhat revealing. As expected, in the absence of protection mechanisms, GLAs are capable of reconstructing recognizable images from intercepted gradients. The application of DP-SGD proves, however, to be a highly effective countermeasure, successfully preventing reconstruction and resulting in outputs that amount to little more than noise, with very low structural similarity scores.

By contrast, the most surprising and \textit{concerning} result comes from the evaluation of PDP-SGD in the context of GLAs. Despite its formal differential privacy guarantees, PDP-SGD does \textit{not} provide \textit{discernible protection} against this type of attack in our experiments. Reconstructed images under PDP-SGD were practically indistinguishable from those obtained without protection, with SSIM values close to the \textit{non-privacy} scenario. This suggests that, although a mechanism may satisfy the theoretical definition of $(\varepsilon, \delta)$-DP, the specific way in which privacy is introduced (e.g., parameter-proportional noise instead of \textit{clipping} and noise addition) may not suffice to obfuscate the information exploited by certain attacks.

Ultimately, we emphasize the importance of not relying solely on theoretical privacy guarantees, but also on \emph{empirically evaluating} the resilience of DP mechanisms against attack vectors that are relevant and specific to the application setting. The apparent disconnect between the formal protection of PDP-SGD and its practical ineffectiveness against GLAs, in our view, deserves deeper investigation.

\section*{Acknowledgement}

We gratefully acknowledge the support of the Ministry of
Economy, Industry, and Competitiveness of Spain under Grant
No. INCEPTION (PID2021-128969OB-I00), as well as the Basque
Government under the grant DEUSTEK5 -- Human-Centric Computing for Smart Sustainable Communities and Environments (IT1582-22).

\bibliography{bibliography.bib}
\bibliographystyle{icml2025}

\newpage
\appendix
\onecolumn
\section{Reconstruction Attack via Gradient Leakage}

\subsection{Attack in a Non-Differentially Private Setting}

\begin{figure*}[!h]
    \centering
    \includegraphics[width=\linewidth]{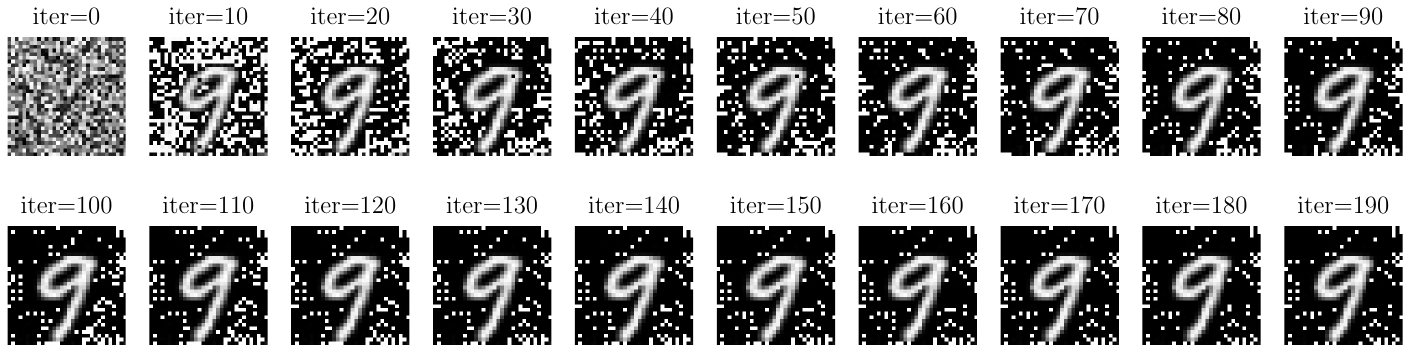}
    \caption{Reconstruction Progression of a Training Example from Model Gradients (\textbf{No Privacy})}
\end{figure*}

\subsection{Attack in a Differentially Private Setting (DP-SGD)}

\begin{figure*}[!h]
    \centering
    \includegraphics[width=\linewidth]{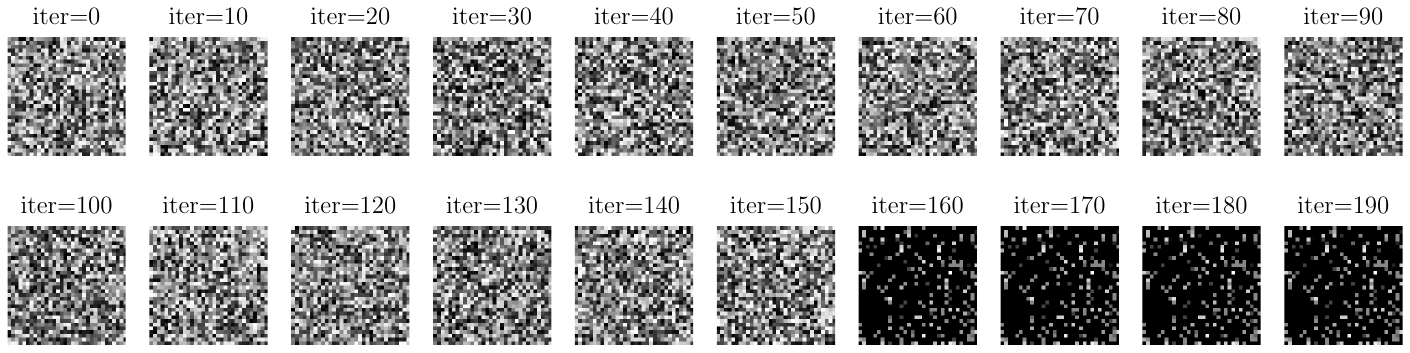}
    \caption{Reconstruction Progression of a Training Example from Model Gradients (\textbf{DP-SGD})}
\end{figure*}

\subsection{Attack in a Differentially Private Setting (PDP-SGD)}

\begin{figure*}[!h]
    \centering
    \includegraphics[width=\linewidth]{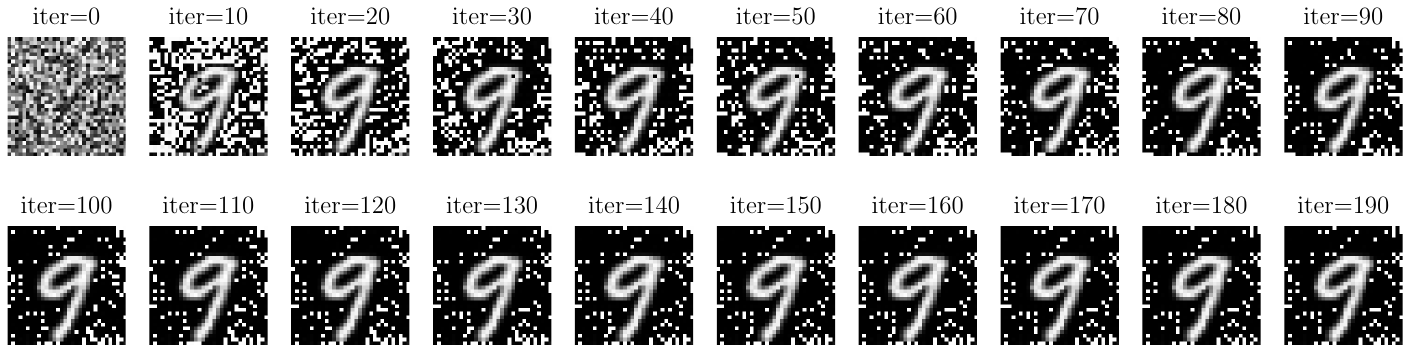}
    \caption{Reconstruction Progression of a Training Example from Model Gradients (\textbf{PDP-SGD})}
\end{figure*}

\end{document}